\documentclass[10pt,a4paper,english,3p,number,sort&compress]{elsarticle}
\usepackage{algorithm}
\usepackage{algpseudocode}
\usepackage{pifont}
\usepackage{amsmath}
\usepackage{amssymb}
\usepackage{comment}
\usepackage{amsfonts}
\usepackage{graphicx}
\usepackage{hyperref}
\usepackage{float}
\usepackage{color}
\usepackage{subcaption}
\usepackage{amsthm}
\usepackage{hyperref}

\journal{journal}

\begin{document}

\begin{frontmatter}

\title{Enhancing Continuous Time Series Modelling with a Latent ODE-LSTM Approach}

\author[inst1]{C. Coelho\corref{cor1}}
\ead{cmartins@cmat.uminho.pt}
\author[inst1]{M. Fernanda P. Costa}
\ead{mfc@math.uminho.pt}
\author[inst1,inst2]{L.L. Ferrás}
\ead{lferras@fe.up.pt}

\cortext[cor1]{Corresponding author} 

\affiliation[inst1]{organization={Centre of Mathematics (CMAT)},
            addressline={University of Minho}, 
            city={Braga},
            postcode={4710 - 057}, 
            country={Portugal}}

\affiliation[inst2]{organization={Department of Mechanical Engineering - Section of Mathematics},
            addressline={University of Porto}, 
            city={ Porto}, 
           postcode={4200-465}, 
            country={Portugal}}

\begin{abstract}

Due to their dynamic properties such as irregular sampling rate and high-frequency sampling, Continuous Time Series (CTS) are found in many applications.
Since CTS with irregular sampling rate are difficult to model with standard Recurrent Neural Networks (RNNs), RNNs have been generalised to have continuous-time hidden dynamics defined by a Neural Ordinary Differential Equation (Neural ODE), leading to the ODE-RNN model.
Another approach that provides a better modelling is that of the Latent ODE model, which constructs a continuous-time model where a latent state is defined at all times. The Latent ODE model uses a standard RNN as the encoder and a Neural ODE as the decoder. However, since the RNN encoder leads to difficulties with missing data and ill- defined latent variables, a Latent ODE-RNN model has recently been proposed that uses a ODE-RNN model as the encoder instead.

Both the Latent ODE and Latent ODE-RNN models are difficult to train due to the vanishing and exploding gradients problem. To overcome this problem, the main contribution of this paper is to propose and illustrate a new model based on a new Latent ODE using an ODE-LSTM (Long Short-Term Memory) network as an encoder - the Latent ODE-LSTM model. To limit the growth of the gradients the Norm Gradient Clipping strategy was embedded on the Latent ODE-LSTM model.

The performance evaluation of the new Latent ODE-LSTM (with and without Norm Gradient Clipping) for modelling CTS with regular and irregular sampling rates is then demonstrated. Numerical experiments show that the new Latent ODE-LSTM performs better than Latent ODE-RNNs and can avoid the vanishing and exploding gradients during training.

Code implementations developed in this work are available at \href{https://github.com/CeciliaCoelho/LatentODELSTM}{github.com/CeciliaCoelho/LatentODELSTM}.

\end{abstract}


\end{frontmatter}

\section{Introduction}

Feed-forward Neural Networks (NNs) propagate information in a unidirectional manner, moving from the input layer through the hidden layers, until the output layer is reached. In contrast, RNNs \cite{elmanFindingStructureTime1990} feature a feedback mechanism between two or more layers, making RNNs ideal for modelling and processing sequential data, such as time series data.


However, RNNs can only process regularly sampled data \cite{chenNeuralOrdinaryDifferential2019,rubanovaLatentOrdinaryDifferential2019}, but real world data sequences are often sampled irregularly. To mitigate this problem, the irregularly-sampled data are rewritten as regularly-sample data, by dividing the time interval into equally-sized intervals, and assigning or aggregating observations using averages.
This type of preprocessing can destroy information, in particular about measurement time, and can also lead to an extra source of error \cite{chenNeuralOrdinaryDifferential2019, rubanovaLatentOrdinaryDifferential2019}. Moreover, RNNs process continuous time series as discrete-time sequence data, complicating real-time processing. In general, RNNs are only suitable for processing moderate-length regular sequence data, with few missing values and small time intervals between observations.

In \cite{chenNeuralOrdinaryDifferential2019}, the authors propose a different solution for handling time series using deep learning - Neural Ordinary Differential Equations (Neural ODEs). In Neural ODEs, the iterative updates of hidden states of the RNN are seen as an Euler discretization of a continuous transformation. Thus, instead of specifying a discrete sequence of hidden layers, it parameterises the derivative of the hidden state using a neural network.
This continuously defined dynamics can naturally incorporate data which arrives at arbitrary times (irregularly sampled data).

\smallskip
Variational Autoencoders (VAEs) are generative models that learn a distribution over data. They provide better predictive accuracy when few data is available, better extrapolation for long time horizons, and the possibility of generating new samples from the original data \cite{kingmaIntroductionVariationalAutoencoders2019}.
In \cite{chenNeuralOrdinaryDifferential2019} the authors propose a VAE with a RNN encoder and a Neural ODE decoder, for time series data.
This new architecture is known as Latent ODE, and improves the performance of the VAEs by representing each time series by a continuous latent trajectory that allows for forward and backward extrapolations in time. The Latent ODE was further improved by Rubanova et al. 
\cite{rubanovaLatentOrdinaryDifferential2019}, where the authors proposed a modification to the encoder of a Latent ODE so that the state transitions of the RNN are defined by a Neural ODE, ODE-RNN, taking advantage of the information given by the sampling intervals of the data. This new architecture was designated by Latent ODE-RNN.

It is known from the literature that RNNs can suffer from the problem of vanishing and exploding gradients \cite{bengioLearningLongtermDependencies1994}, making these networks difficult to train.
To mitigate this problem, the Long Short-Term Memory (LSTM) network \cite{hochreiterLongShorttermMemory1997} was developed, and in \cite{lechnerLearningLongTermDependencies2020} the authors proposed an ODE-LSTM to overcome the vanishing and exploding gradients problem of ODE-RNNs.

In this work, we prove that Latent ODE-RNNs still suffer from the vanishing and exploding gradients problem and propose replacing the ODE-RNN encoder by an ODE-LSTM, thus avoiding gradient dissipation and improving the performance of the network when learning long-term dependencies.

Since the LSTM networks are still prone to gradient explosion \cite{sutskeverSequenceSequenceLearning2014} \cite{pascanuDifficultyTrainingRecurrent2013}, we propose combining the Latent ODE-LSTM architecture with norm gradient clipping, a technique used to control gradients by rescaling \cite{pascanuDifficultyTrainingRecurrent2013}, which is referred to as Latent ODE-LSTM and Latent ODE-LSTM+gradient clipping throughout the paper. Table \ref{table:LatentODE evolution} shows the Encoder and Decoder used in each variant of the Latent ODE architecture.
We compare the newly proposed Latent ODE-LSTM (with and without gradient clipping) to a Latent ODE-RNN using synthetic irregularly sampled, gradually sparser time series, and real-life regularly and irregularly sampled time series. The results show that the new architecture outperforms Latent ODE-RNN, and consequently the classical architectures.

\begin{table}[H]
\resizebox{\textwidth}{!}{
\begin{tabular}{lll}
\hline
Encoder-decoder models & Encoder  & Decoder \\ \hline
Latent ODE \cite{chenNeuralOrdinaryDifferential2019}             & RNN      & ODE     \\
Latent ODE-RNN \cite{rubanovaLatentOrdinaryDifferential2019}        & ODE-RNN  & ODE     \\
\textbf{Latent ODE-LSTM (ours)}       & \textbf{ODE-LSTM}       & \textbf{ODE}     \\ 
\textbf{Latent ODE-LSTM+gradient clipping (ours)}       & \textbf{ODE-LSTM}       & \textbf{ODE}     \\ \hline
\end{tabular}
}
\caption{Evolution of VAEs with a Neural ODE decoder - Latent ODEs family. Over time, VAEs architectures have been proposed, using different encoders to solve the difficulties of the previous versions.}
\label{table:LatentODE evolution}
\end{table}

The paper is organised as follows. Section \ref{sec:background} presents a brief review of essential concepts such as RNNs, LSTM networks, Norm Gradient Clipping,  Neural ODEs, Autoencoders, VAEs,  Latent ODEs, Latent ODE-RNN.

Section \ref{sec:method} is devoted to the new Latent ODE-LSTM architecture. It is shown that the vanishing gradient problem is mitigated using the LSTM. Also, the exploding gradient problem is addressed by combining the Latent ODE-LSTM with norm gradient clipping (this section relies on \ref{sec:gradient} where we prove that RNNs suffer from the vanishing and exploding gradients problem \cite{bengioLearningLongtermDependencies1994,pascanuDifficultyTrainingRecurrent2013}, and, building on that, we also prove that Latent ODEs and Latent ODE-RNNs suffer from the same problem).

In Section \ref{sec:experiments} we evaluate the performance of the different architectures by considering the reconstruction and extrapolation of spirals, and numerical experiments with two real-life datasets (one with regularly sampled data and the other with irregularly sampled data).  The paper ends with the conclusions and future work in Section \ref{sec:conclusion}.

\section{Background} \label{sec:background}
This section provides some of the background information needed for the next sections.

Let $\mathcal{X}=(x_1,x_2,\dots,x_N)$ be an input sequential data of length $N$, with $x_i \in  \mathbb{R}^d$ denoting the input at time step $i$ ($i=1,\dots,N$). Let $\mathcal{Y}=(y_1,y_2,\dots,y_N)$ be the desired response sequential data, with $y_i \in  \mathbb{R}^p$ 
denoting the response vector at time step $i$, and, let  $\mathcal{\hat{Y}}=(\hat{y}_1,\hat{y}_2,\dots,\hat{y}_N)$ be the output response sequential data produced by an architecture, with $\hat{y}_i \in  \mathbb{R}^p$ denoting the output response vector at time step $i$. 

\subsection{RNNs}

There are a number of works in the literature, \cite{lawrenceUsingNeuralNetworks1997,vaizHybridModelForecast2016}, in which NNs are used to handle sequential data. However, when using sequential data $\mathcal{X}$, NNs only provide \emph{independent} data values, and there is no way to convey a dependency or ordering idea between each value of the data sequence. Furthermore, because NNs have a fixed number of neurons in the input layer, it is not possible to generalise the model to input and output sequences with arbitrary lengths that were not used during training \cite{waibelPhonemeRecognitionUsing1989}. 

To address these issues, RNNs were developed \cite{elmanFindingStructureTime1990}.  RNNs have feedback loops that allow multiple input values to be fed sequentially, enabling sequential data modelling \cite{elmanFindingStructureTime1990}. 

\begin{figure}[H]
    \centering
    \resizebox{\textwidth}{!}{
    \includegraphics{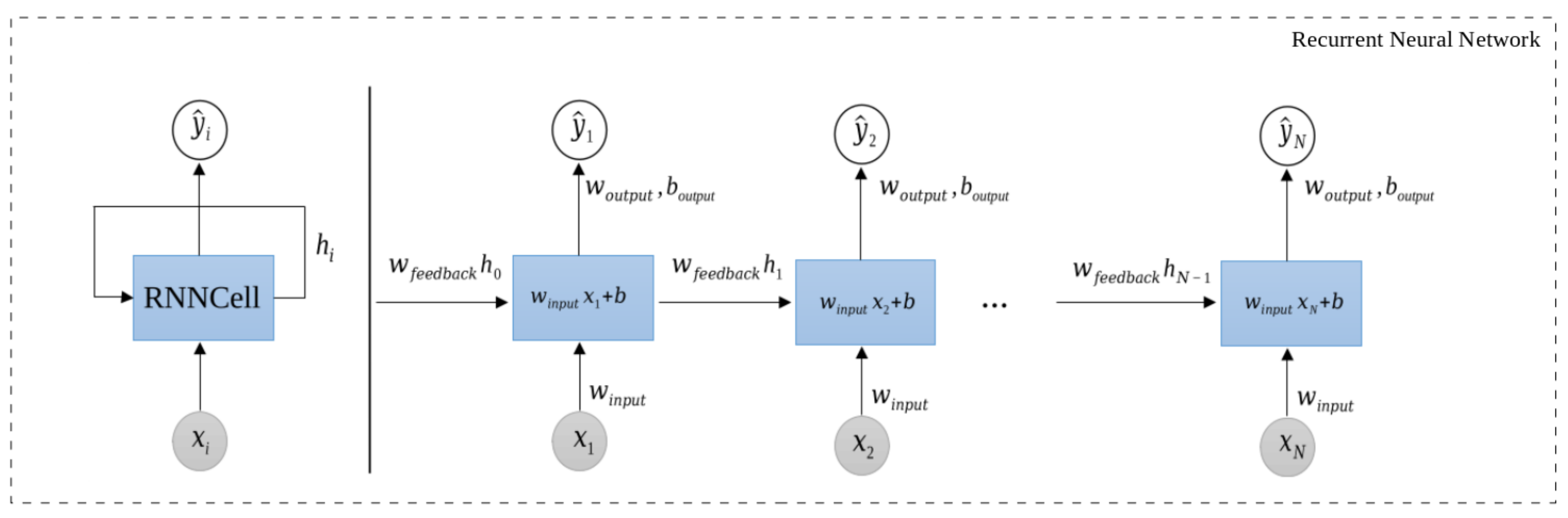}
    }
    \caption{Schematic representation of a RNN applied to a sequential data of arbitrary length. The feedback loop can be unrolled over time and represented as $N$ copies of the RNN cell, resembling a deep fully-connected feed-forward NN with a depth of $N$.}
    \label{fig:RNN}
\end{figure}

An RNN builds a sequence of $n$-dimensional hidden state vectors $h_i \in \mathbb{R}^n$, where $h_i$ catches the essential characteristics of the input sequences from the first time step to $i$. The left side of Figure~\ref{fig:RNN} shows a simple RNN, with a hidden layer $h_i$ and its feedback loop. A RNN unrolling through time, which is the same hidden layer represented once per time step $i$ ($i=1,\dots,N$) is represented on the right side of Figure~\ref{fig:RNN}. Therefore, a RNN is a deep feed-forward NN with $N$ layers, where each layer has a number of neurons, $n$, equal to the length of $h_i$.   

 At time step $i$, the hidden state $h_i$ depends on the input vector $x_i$ and the previous hidden state $h_{i-1}$ from time step $i-1$. The RNN cell that constitutes the layers of a RNN is defined by: 
\begin{equation}
\label{eq:bck_rnn1}
h_{i} = \sigma(\textbf{w}_{\text{feedback}} h_{i-1} +  \textbf{w}_{\text{input}}x_{i} +b)
\end{equation}
Here, $\sigma$ is an activation function, and the initial hidden state vector $h_0$ needs to be initialised. 
The matrix $\textbf{w}_{\text{input}} \in \mathbb{R}^{n \times d}$, contains the weights that link the input and hidden state vectors. 
The matrix $\textbf{w}_{\text{feedback}} \in \mathbb{R}^{n \times n}$ contains the weights that link two hidden state vectors at time step $i-1$ and $i$, and $b \in \mathbb{R}^n$ the bias vector of the current hidden state. 

At time step $i$, having the hidden state vector $h_i$, the output vector $\hat{y}_i$ is computed by: 

\begin{equation}
\label{eq:outp_rnn1}
\hat{y}_{i} = \sigma(\textbf{w}_{\text{output}}h_{i} + b_{\text{output}})
\end{equation}
where the matrix $\textbf{w}_{\text{output}} \in \mathbb{R}^{p \times n}$ contains the weights that link the hidden state and the output vectors, with bias vector $b_{\text{output}}$.

In a RNN, the weight matrices and bias vectors are constant across all of time steps $i$, that is, all parameters $\theta: \textbf{w}_{\text{input}},\, \textbf{w}_{\text{feedback}},\, b, \textbf{w}_{\text{output}},\, b_{\text{output}}$ are shared across the time steps. 
The strategy of \emph{parameter sharing} is noteworthy as it enables a significant reduction in the number of parameters that an RNN needs to learn. This is done by assuming that the shared parameters can capture all the essential sequential features.

Note that $h_i$ is a function of $x_i$ and $h_{i-1}$, which is a function of $x_{i-1}$  and $h_{i-2}$, which is a function of $x_{i-2}$ and $h_{i-3}$, etc. Thus,  $h_i$ is a function of all the inputs since the \emph{instant} $i=1$: 
\begin{equation}
h_{i} = \sigma(\textbf{w}_{\text{feedback}}(...\,  \overbrace{\sigma(\textbf{w}_{\text{feedback}} \underbrace{\sigma( \textbf{w}_{\text{feedback}}h_0+  \textbf{w}_{\text{input}}x_1  + b)}_{h_1}  +   \textbf{w}_{\text{input}}x_2 + b)}^{h_2}  + ...))  +  \textbf{w}_{\text{input}}x_{i}  + b
\end{equation}
with $h_0$ usually initialised as the null vector. \\
Since the output $\hat{y}_i$ and hidden state $h_i$ are functions of all the inputs from previous time steps, it is considered to exist a form of \emph{memory}, being the RNN cell also known as \emph{short-term memory cell}.

The update of formula \ref{eq:bck_rnn1} performed by an RNN cell is often represented by $h_i = RNNCell(h_{i-1}, x_i)$. 

To train a RNN, the RNN is unrolled through time and the BackPropagation Through Time (BPTT) technique is applied. 

However, when training very deep networks, such as RNNs for long sequences, the vanishing or exploding gradients problem may arise \cite{bengioLearningLongtermDependencies1994} \cite{pascanuDifficultyTrainingRecurrent2013}. Moreover, RNNs are only suitable for handling regularly-sampled data with moderate length since, with the increase in length, the first time steps of the sequence will progressively be forgotten.

We proved in \ref{proof:RNN}, that RNNs suffer from the vanishing and exploding gradients problem.

\subsection*{LSTM Networks} 

To alleviate the vanishing gradient problem and the short-term memory problem in RNNs, LSTM networks have been proposed
\cite{hochreiterLongShorttermMemory1997}. 
Like RNNs, LSTMs use feedback loops but with an additional internal cell state $C_i  \in \mathbb{R}^n$, which corresponds to long-term memory, and three differentiable gates (input gate vector $I_i  \in \mathbb{R}^n$, forget gate vector $F_i  \in \mathbb{R}^n$, output gate vector $O_i \in \mathbb{R}^n$), to control the states $h_i$ and $C_i$. 

At time step $i$, the three gate vectors are updated accordingly:
\begin{equation}\label{eq:gates}
\begin{array}{rcl}
   I_{i}&=  &   \sigma( \textbf{w}_{xin}x_i +  \textbf{w}_{hin}h_{i-1}+b_{in})\\
   F_{i}&= &\sigma(  \textbf{w}_{xf}x_i+ \textbf{w}_{hf}h_{i-1} +b_f) \\
    O_i&=&\sigma( \textbf{w}_{xo}x_i +  \textbf{w}_{ho}h_{i-1}+b_o)
\end{array}
\end{equation}
Here, $\sigma$ is the sigmoid activation function. Each gate is a function of the input vector $x_i$ and the previous time step hidden state $h_{i-1}$. Each gate has a weight matrix from input to gate vector, and from the hidden state vector to the gate vector, and the respective bias vector, with $\textbf{w}_{xin}, \textbf{w}_{xf},  \textbf{w}_{xo} \in \mathbb{R}^{n \times d}$, $\textbf{w}_{hin}, \textbf{w}_{hf},  \textbf{w}_{ho} \in \mathbb{R}^{n \times n}$ and $b_{in},  b_{f}, b_{o}, \in \mathbb{R}^{n}$. 

Each gate in an LSTM network has a distinct role. The input gate vector $I_i$ allows the NN to decide how much of the input vector, via the memory state $\tilde{C_i} \in \mathbb{R}^n$, is allowed to influence the memory state $C_i$. The forget gate vector $F_i$ allows the NN to decide how much of the previous memory vector $C_{i-1}$ should be forgotten,  and the output gate vector $O_i$ allows the NN to decide how much of the internal memory state $C_i$ should be retained for the hidden state $h_i$.
At time step $i$, given the current input $x_i$ and the previous hidden state $h_{i-1}$, LSTM first calculates the candidate memory update $\tilde{C}_i$ as follows:
\begin{equation}
\tilde{C}_{i}= \tanh(\textbf{w}_{xc} x_i  + \textbf{w}_{hc} h_{i-1} +b_c).
\end{equation}
Here, $\tanh$ is the hyperbolic tangent activation function. The matrix $\textbf{w}_{xc} \in \mathbb{R}^{n \times d}$ contains the weights between the input vector and the candidate memory cell, $\textbf{w}_{hc} \in \mathbb{R}^{n \times n}$ contains the weights between the hidden state and the candidate memory cell, and $b_c \in \mathbb{R}^{n}$ is the corresponding bias vector. 

Then, the three gates are used to calculate the vectors of the internal memory $C_i$ and hidden state $h_i$:
\begin{equation}\label{eq:lstm}
\begin{array}{rcl}
     C_{i}&=& F_i \odot C_{i-1}+I_i \odot \tilde{C}_i  \\
     h_{i}&=& O_{i} \odot \tanh(C_i),
\end{array}
\end{equation}

\noindent where $\odot$ is the element-wise product operation.

Finally, as in RNNs, the output vector $\hat{y}_i$ at time $i$ is computed using the hidden state vector $h_i$, as follows:
\begin{equation}
\nonumber
\hat{y}_{i} = \sigma(\textbf{w}_{\text{output}}h_{i} + b_{\text{output}}).
\end{equation}

\begin{figure}[H]
    \centering
    \includegraphics[scale=0.4]{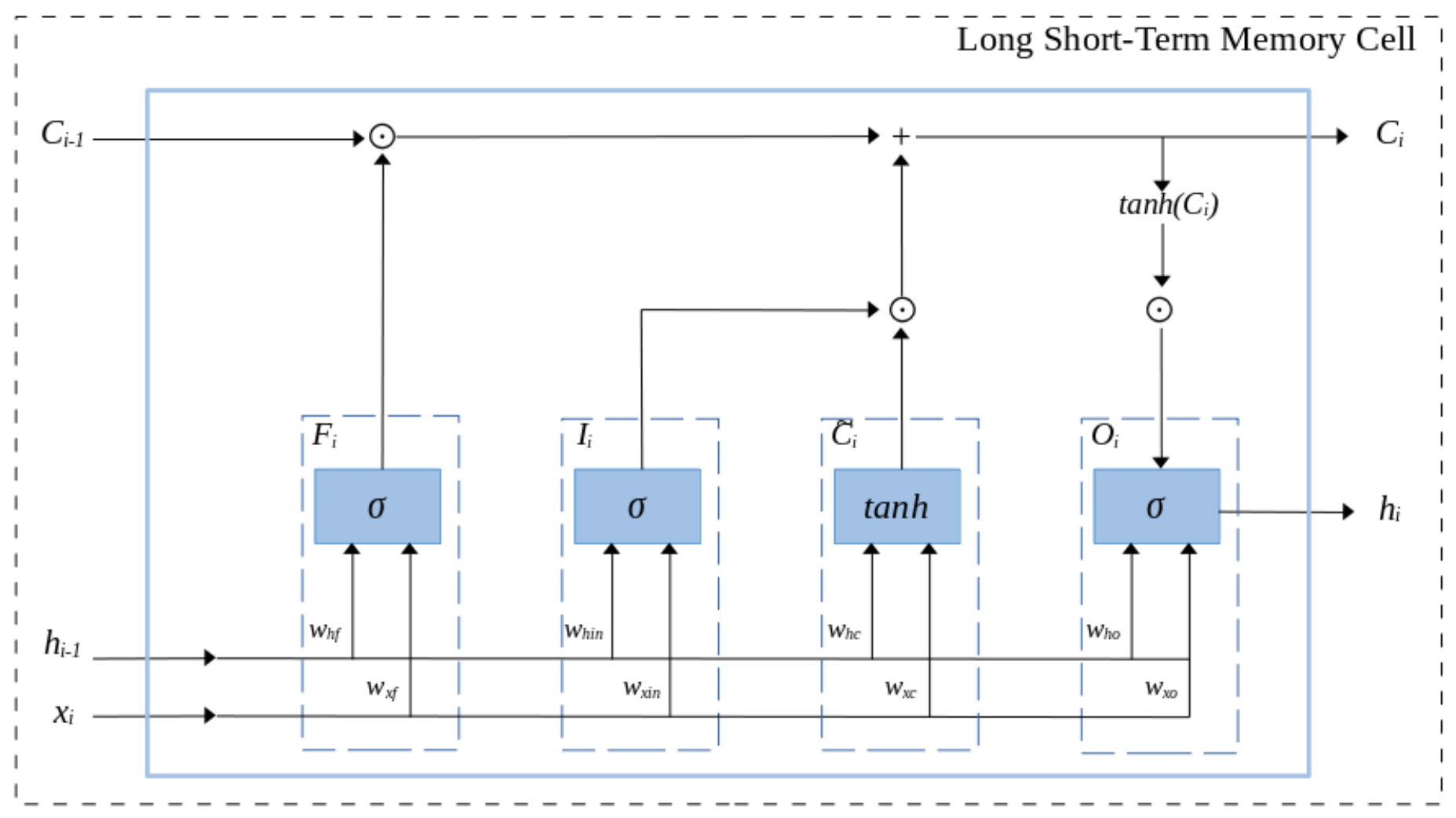}
    \caption{Long Short-Term Memory cell scheme.}
    
    \label{fig:LSTM}
\end{figure}

\noindent Here, $\sigma$ is the sigmoid activation function, $\textbf{w}_{\text{output}} \in \mathbb{R}^{p \times n}$ is the weight matrix that contains the weights that link the hidden state and the output vectors, and $b_{\text{output}}$ the bias vector. The initial state vector $h_0$ and the initial internal memory state vector $C_0$ are initialised, usually, as the null vector.  Figure \ref{fig:LSTM} shows the scheme of a LSTM cell.

The update of formulas \eqref{eq:gates}-\eqref{eq:lstm} performed by an LSTM cell are often represented by\\ $(C_i,h_i) = LSTMCell(C_{i-1},h_{i-1},x_i)$.

To train a LSTM network, like in a RNN, it is unrolled through time, and a modified version of the BPTT technique to deal with element-wise operation in the update formula \eqref{eq:lstm}, is applied. 

Due to their design, LSTMs are suitable for handling regularly-sampled long discrete sequence data. Moreover, they ensure constant error flow through the network due to the dependence of the previous memory cell $C_{j-1}$ with the current $C_j$, preventing gradients from vanishing \cite{hochreiterLongShorttermMemory1997}.

Note that, the computational cost of a LSTM is higher than for a RNN model due to the more complex architecture with multiple gates and memory cells. This increased complexity generally requires more computations during both training and prediction. 
Despite the increase of computational cost, LSTMs are an attractive alternative when favouring accuracy over computational cost.

\subsection*{Neural ODE}

The authors in \cite{chenNeuralOrdinaryDifferential2019} observed that some neural network architectures, such as RNN and LSTM, learn how one hidden state, $h_{t}$, differs from the next one, $h_{t+1}$, resembling an Euler discretisation,

 \begin{equation}
h_{t+1} = h_{t} + f(h_{t}, \theta)
\end{equation}

\noindent where $\theta=(\boldsymbol{w}, b)$ and $t = 0, \dots, N$. When the steps from $h_{t}$ to $h_{t+1}$ are infinitesimally small, these computations resemble a continuous dynamic process \cite{chenNeuralOrdinaryDifferential2019}.
Thus, the authors in \cite{chenNeuralOrdinaryDifferential2019} proposed Neural ODEs, a NN that models an ODE to the hidden states dynamics:

\begin{equation}
\dfrac{dh(t)}{dt}=f(h(t),t,\theta),\,\,\text{with}\,\,h(0) = h_0.
\label{eq:NODE4}
\end{equation}

\begin{figure}[H]
    \centering
    \resizebox{\textwidth}{!}{
    \includegraphics{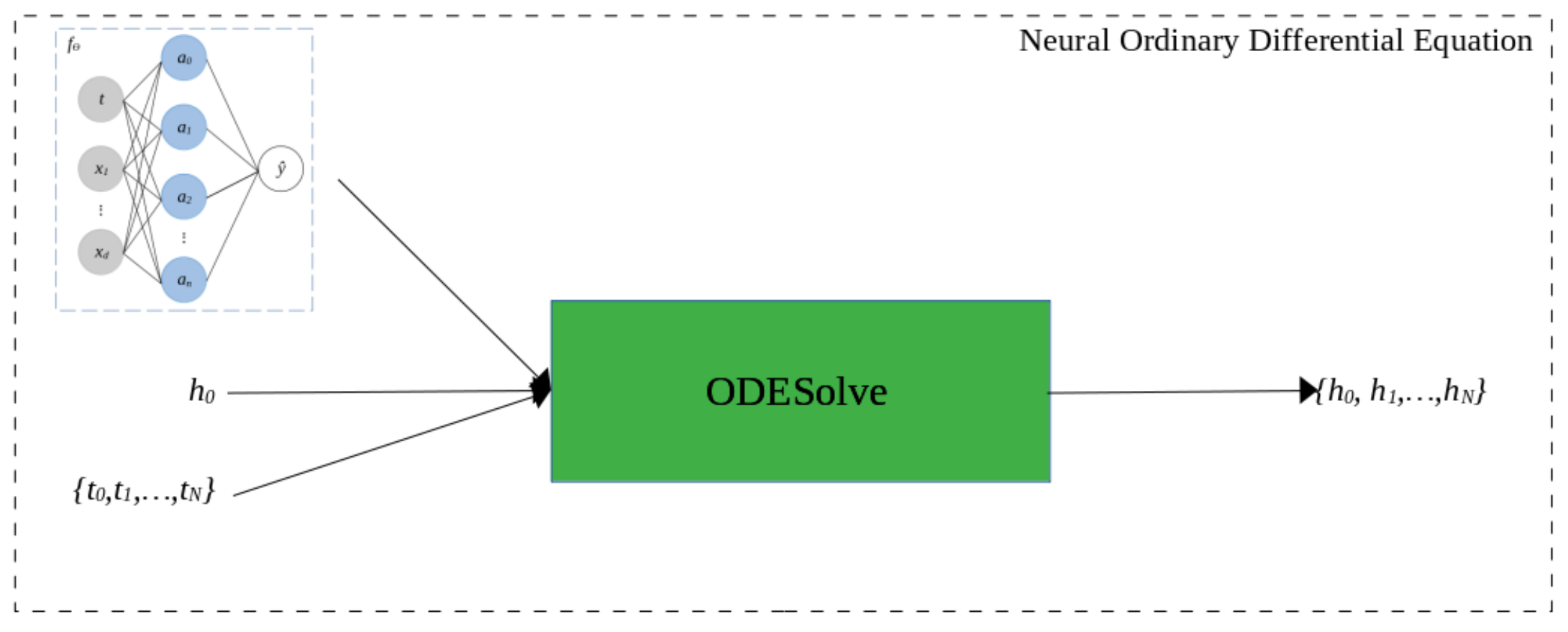}
    }
    \caption{Architecture of a Neural Ordinary Differential Equation. In the training phase, the time interval given to the ODE solver is $[t_0, t_N]$ and in the testing phase, predictions can be made in an arbitrary time interval $[t_0, t_f]$.}
    \label{fig:Neural ODE}
\end{figure}

A Neural ODE is composed of two parts, a NN that models a function dynamics $f_\theta$ by optimising the parameters $\theta$ and an ODE solver. The final result of training a Neural ODE is the function $f_{\theta}$. Then, to make predictions, the solution of the ODE initial value problem \eqref{eq:NODE4}, is computed for a given time interval ($t_{0}, t_f$) by an ODE solver:

\begin{equation*}
\{h_t\}_{t=0}^{N} = ODESolve(f_{\theta}, h_0, (t_0,t_f)).
\label{eq:NODE3}
\end{equation*}
Figure \ref{fig:Neural ODE} shows the architecture of the Neural ODE.

Note that during training, to optimise the parameters $\theta$, backpropagating through the ODE solver can be used, but it has a high memory cost and introduces additional numerical errors \cite{chenNeuralOrdinaryDifferential2019}. Thus, the authors in \cite{chenNeuralOrdinaryDifferential2019,pontryaginMathematicalTheoryOptimal2018} propose to use the Adjoint Sensitivity method to train a neural ODE. It scales linearly with the problem size, has low memory cost and explicitly controls the numerical error
\cite{chenNeuralOrdinaryDifferential2019,pontryaginMathematicalTheoryOptimal2018}.

\subsection*{Autoencoders}

Autoencoders are an unsupervised learning process with the goal of reducing the dimension of input data, \textit{i.e}, it is a form of data compression \cite{Goodfellow-et-al-2016}. Through the process of compressing the input data, encoding it and then reconstructing it as output, autoencoders allow one to reduce dimensionality and focus only on the areas that are truly valuable. They have application in information retrieval, anomaly detection, image processing (image denoising as well as super-resolution), etc.

\begin{figure}[H]
    \centering
    \includegraphics[scale=0.4]{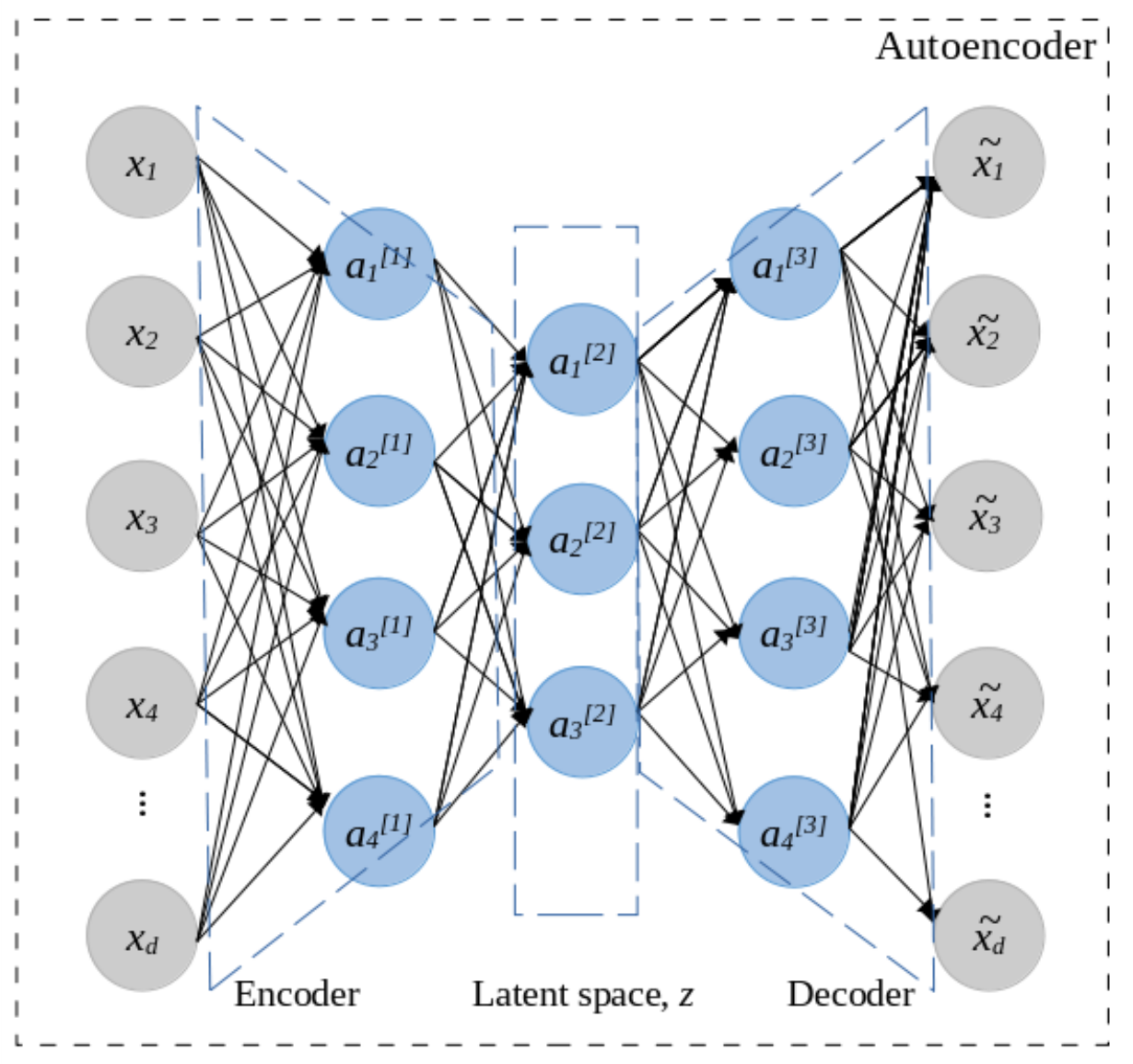}
    \caption{Architecture of an Autoencoder consisting of an encoder and a decoder.}
    \label{fig:autoencoder}
\end{figure}

The architecture of an autoencoder is very simple and information only flows in one way, forward, meaning that we have a feed-forward NN. This network is usually treated as being composed of two main components (that can be seen as separate NNs and that have different architectures), an encoder and a decoder, which learn an encoding-decoding scheme that minimises the loss using an iterative optimisation process \cite{Goodfellow-et-al-2016}.
The encoder learns a mapping from the data, $x \in \mathbb{R}^d$, to a low-dimensional latent space with dimension $l$, $z \in \mathbb{R}^l, \,\,\, l<d$, leading to a bottleneck (layer with the fewest number of neurons in the network), while the decoder learns a mapping from the latent space, $z$, to a reconstructed observation, $\tilde{x} \in \mathbb{R}^d$ with $\tilde{x} \approx x$ \cite{Goodfellow-et-al-2016}. Figure \ref{fig:autoencoder} shows the architecture of an autoencoder.

The goal is to train the model to reconstruct the original data as better as possible, so that the cost function $\mathcal{L}_{\theta}$ minimises the difference between the original input $x$ and the reconstruction $\tilde{x}$, by optimising the network parameters $\theta$ using the backpropagation method \cite{Goodfellow-et-al-2016}.

The main difference between autoencoders and traditional NNs is that instead of training an input/output pair $(x,y)$ and modelling a transformation $x \mapsto y$, autoencoders try to fit the input data to themselves $x \mapsto x$.

\subsection*{VAEs}

 Autoencoders attempt to reconstruct a given input $x$ by learning mappings to and from a lower-dimensional vector $z$. However, an autoencoder is not a generative model as it does not model a distribution of the data, being unable to sample new data from a learned distribution.

Variational Autoencoders \cite{kingmaAutoEncodingVariationalBayes2014} (VAEs) on the other hand, are generative models in which samples are generated by a neural network $NN_\alpha$ applied to a random latent variable $z$. This latent variable is drawn from a distribution with all possible values, called the prior distribution $p_\alpha(z)$ \cite{kingmaIntroductionVariationalAutoencoders2019},

\begin{equation*}
x = NN_\alpha(z).
\end{equation*}

Unlike autoencoders, VAEs consist of probabilistic versions of encoders and decoders.
The goal of VAEs is to find the true posterior distribution $p_\alpha(z|x)$ learned by the encoder so that latent variables can be inferred from data, and to find a stochastic decoder $p_\alpha(x|z)$ so that a sample can be reconstructed from a random latent variable $z$, \cite{kingmaIntroductionVariationalAutoencoders2019}.

\begin{figure}[H]
    \centering
    \resizebox{\textwidth}{!}{
    \includegraphics[]{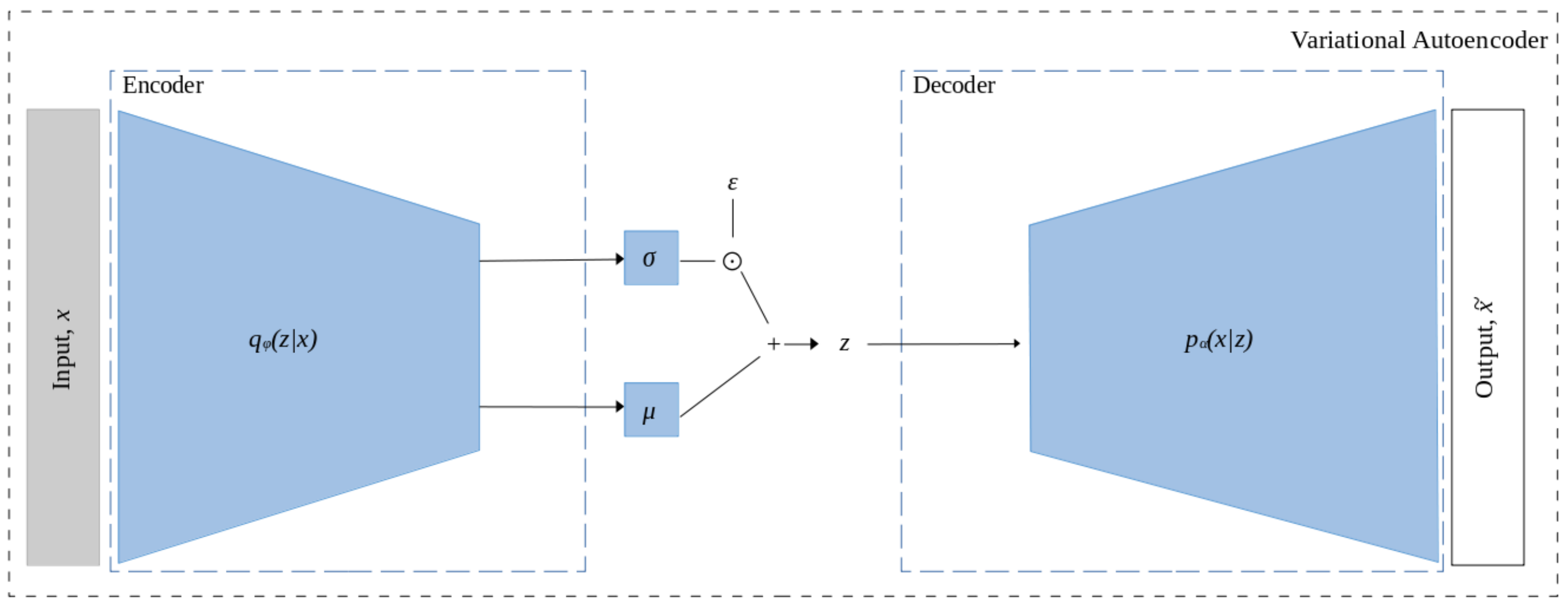}
    }
    \caption{Architecture of a VAE consisting of two main components: a probabilistic encoder and a probabilistic decoder.}
    \label{fig:VAE}
\end{figure}

However, computing the true posterior distribution, $p_\alpha(z|x)$, is difficult because it is too expensive due to the enumeration of all values of $x$. To overcome this issue, the authors in \cite{kingmaAutoEncodingVariationalBayes2014} proposed learning an approximation to the true posterior distribution using a stochastic encoder $q_{\phi}(z|x)$. Thus, given an input $x$  the stochastic encoder generates the corresponding value $z$ (see Figure \ref{fig:VAE})\cite{kingmaIntroductionVariationalAutoencoders2019}.

The mean, $\mu \in \mathbb{R}^l$, and the standard deviation, $\sigma \in \mathbb{R}^l$, are given by the encoder and used to sample the latent variable $z$ given by $z=\mu + \sigma \odot \epsilon$. The generative part of a VAE is created by inducing randomness by adding noise $\epsilon \in \mathbb{R}^l$ to calculate $z$ \cite{kingmaAutoEncodingVariationalBayes2014,kingmaIntroductionVariationalAutoencoders2019}. 

The training of a VAE is done by optimising the parameters of the encoder, $\phi$, and decoder, $\alpha$, simultaneously. This is done by optimising a cost function $\mathcal{L}_\Theta$, with $\Theta=(\phi,\alpha)$, defined as follows: 

\begin{equation*}
    \mathcal{L}_\Theta = \dfrac{1}{N} \sum_{i=1}^{N} l(\tilde{x}_i)
\end{equation*}

\noindent where $l(\tilde{x}_i)$ is the individual cost for each time step. In VAEs, $\mathcal{L}_\Theta$ is the Evidence Lower Bound (ELBO) function \cite{kingmaAutoEncodingVariationalBayes2014}.

\subsection*{Latent ODE}

\begin{figure}[H]
    \centering
    \resizebox{\textwidth}{!}{
    \includegraphics{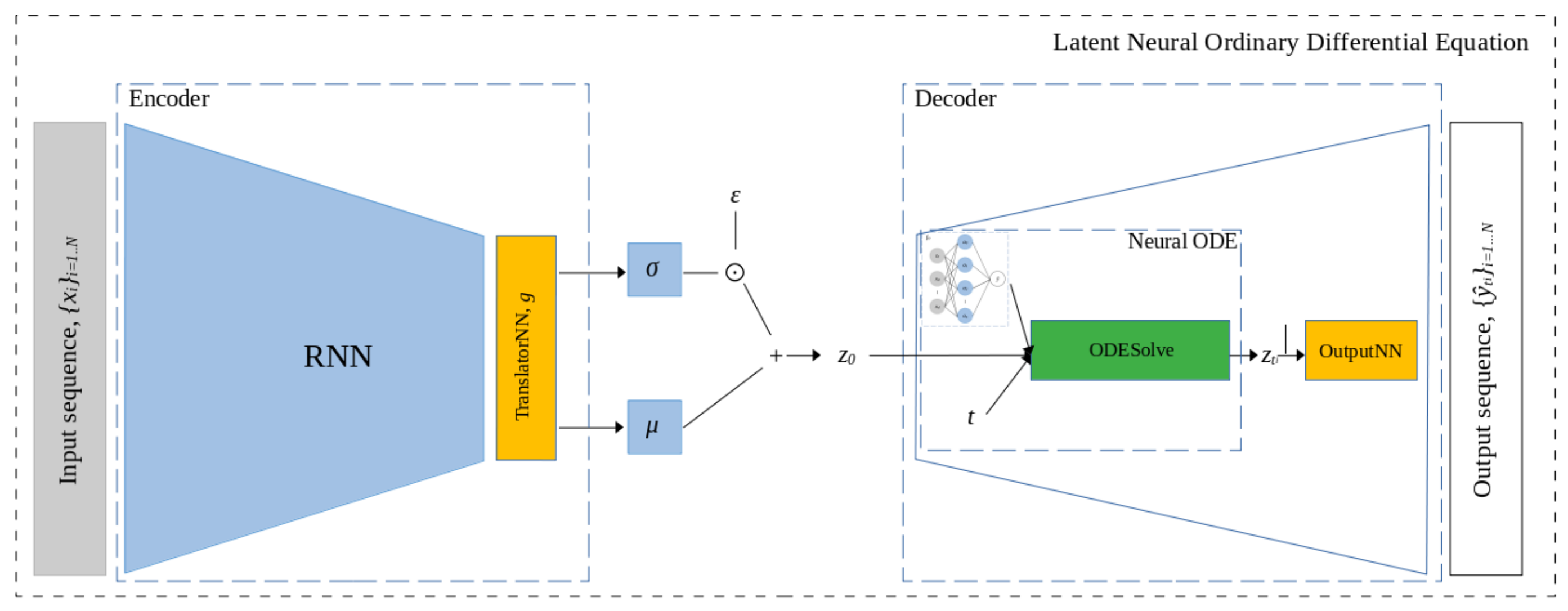}
    }
    \caption{Architecture of a Latent ODE, a VAE composed of a RNN encoder and a Neural ODE decoder.}
    \label{fig:Latent ODE}
\end{figure}

In \cite{chenNeuralOrdinaryDifferential2019} the authors propose Latent ODEs, a VAE architecture using a RNN encoder and a Neural ODE decoder (see Figure \ref{fig:Latent ODE}).

Note that the encoder uses a RNN combined with a translator NN $g$, and the decoder uses a Neural ODE combined with an output NN, that models a distribution $p_\alpha(x|z)$. The RNN output is transformed by $g$ and gives a mean $\mu$ and standard deviation $\sigma$ of a distribution $q_\phi$:

\begin{equation*}
q_{\phi}(z_{0}| \{x_i\}_{i=1}^{N}) = \mathcal{N}(\mu,\sigma) \,\,\, \text{where} \,\,\, \mu,\sigma=g(\text{RNN}(\{x_i\}_{i=1}^{N})).
\end{equation*}

\noindent Therefore, the encoder converts the input sequence $\{x_i\}_{i=1}^N$ into the latent variable of the initial state: $$z_{t_0} \approx z_0=\mu + \sigma \odot \epsilon \in \mathbb{R}^l$$ where $l$ is the dimension of the latent space, $\phi$ are the parameters of the NN encoder and $\epsilon$ is a randomly sampled noise.

Before training (in networks with a Neural ODE decoder), to solve an ODE initial value problem, the sequence must be reversed in time to compute the initial state $z_0$. The initial state $z_0$ is then given to a Neural ODE decoder that learns a continuous latent trajectory $f(z_0,\theta,t)$. Then, an ODE initial value problem is solved using an ODE solver giving the latent variables $z_t$ in the entire time domain $(t_0, \dots, t_N)$:

\begin{equation*}
\{z_{t_i}\}_{i=0}^{N} = ODESolve(f_\theta, z_0, (t_{0}, \dots, t_N)).
\end{equation*}

Finally, a learned distribution reconstructs the sample from the latent space, where $\alpha$ are the parameters of the decoder.

Again, training is done by optimising a cost function $\mathcal{L}_\Theta$, with $\Theta=(\phi,\alpha)$, defined as follows: 

\begin{equation}
\label{eq:loss}
    \mathcal{L}_\Theta = \dfrac{1}{N} \sum_{i=1}^{N} l(\hat{y}_i)
\end{equation}

\noindent where $l(\hat{y}_i)$ is the individual cost for each time step.

We note that the use of a Neural ODE decoder allows easy extrapolation forward or backward in time, since a continuous-time latent trajectory is available. Note that although the latent variable $z_0$ is computed by a stochastic process that introduces randomness into the model, the ODE, which models evolution through time, assumes deterministic dynamics since the initial value problem depends on the initial state $z_0$. Therefore, each state $z_t$ is uniquely defined and dependent on the initial state $z_0$. However, this determinism can be problematic in certain applications that have inherent randomness that cannot be captured by such deterministic models.

We prove in Appendix \ref{proof:Latent ODE}, 
that Latent ODEs suffer from the vanishing and exploding gradients problem due to the RNN in the encoder.

\subsection*{Latent ODE-RNN}

Latent ODEs use a RNN encoder and have difficulty handling irregularly sampled data. In \cite{rubanovaLatentOrdinaryDifferential2019}, the authors propose Latent ODE -RNNs that replace the RNN in Latent ODEs with a ODE-RNN. In this way, the encoder can perceive the sampling time between observations.
The ODE-RNN \cite{rubanovaLatentOrdinaryDifferential2019} is a RNN where the state transitions are defined by a Neural ODE. In this case, the RNN update \eqref{eq:bck_rnn1} is computed with an intermediate hidden state $h'_i \in \mathbb{R}^n$:

\begin{equation}
\label{eq:bckLatentODERNN2}
h_i = RNNCell(h'_i,x_i),
\end{equation}

\noindent  where $h'_i$ is the solution of the ODE initial value problem in the time interval $(t_{i-1},t_i)$ by using the previous hidden state $h_{i-1}$ as the initial value:

\begin{equation}
h'_{i}= ODESolve(f_\theta,h_{i-1},(t_{i-1},t_i)), \,\,\, i=1,\dots,N
\end{equation}

\begin{figure}[H]
    \centering
    \resizebox{\textwidth}{!}{
    \includegraphics{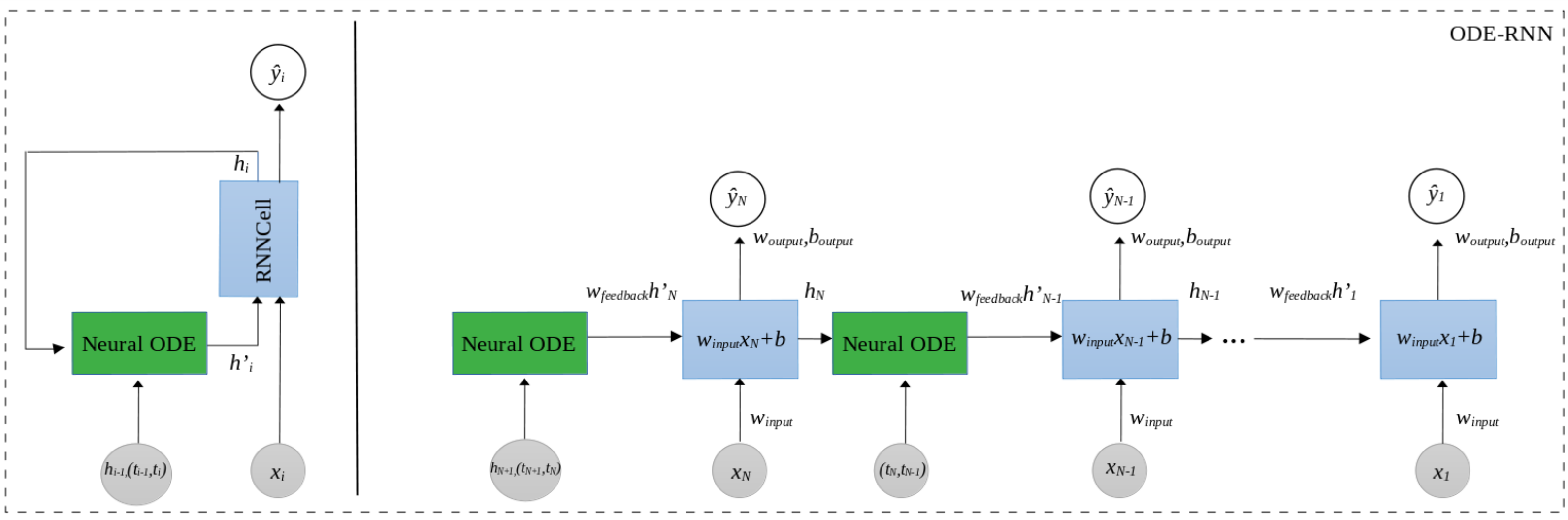}
    }
    \caption{Architecture of an ODE-RNN present in a Latent ODE-RNN encoder.}
    \label{fig:ODE-RNN}
\end{figure}

Figure \ref{fig:ODE-RNN} shows the architecture of an ODE-RNN. Note that the ODE-RNN, in a Latent ODE-RNN, deals with the sequence backwards in time and produces a single output $z'_0$, at time $t_0$:

$$z'_0 = \text{ODE-RNN}_\phi(\{x_i, t_i\}_{i=1}^N).$$

Then, the $z'_0$ is used by the translator network $g$ to output the mean, $\mu$, and standard deviation, $\sigma$: $\mu, \sigma = g(z'_0)$.

The initial value $z_0$ is sampled from $\mathcal{N}(\mu,\sigma)$, $z_0=\mathcal{N}(\mu, \sigma)$, and used as the initial value to compute the solution of the ODE initial value problem, in the decoder.

By replacing the RNN encoder in Latent ODEs, Latent ODE-RNNs were shown to outperform Latent ODEs when learning from irregularly sampled data due to the ability to better learn the approximate posterior distribution \cite{rubanovaLatentOrdinaryDifferential2019}.

However, RNNs suffer from the vanishing and exploding gradients problem \cite{bengioLearningLongtermDependencies1994}. Thus networks such as ODE-RNN, Latent ODE and Latent ODE-RNN, that embed RNNs, suffer from the same problem too. The authors in \cite{lechnerLearningLongTermDependencies2020} proposed ODE-LSTMs that are less prone to this problem.

As proved in Appendix \ref{proof:Latent ODE-RNN}, 
Latent ODE-RNNs suffer from the vanishing and exploding gradients problem due to the RNN in the encoder.

\subsection*{Norm Gradient Clipping}

Norm gradient clipping is a strategy introduced to address the explosion gradient problem by rescaling \cite{pascanuDifficultyTrainingRecurrent2013}. 

If the norm of the gradients is above a predefined threshold, $\lVert \dfrac{\partial \mathcal{L}}{\partial \Theta} \rVert \geq threshold$, then the gradients are updated as follows:
$$\dfrac{\partial \mathcal{L}}{\partial \Theta} = \dfrac{threshold}{\lVert \dfrac{\partial \mathcal{L}}{\partial \Theta} \rVert} \dfrac{\partial \mathcal{L}}{\partial \Theta}$$.

Rescaling is performed after all gradients have been calculated and before the network parameters are updated.
Note that this strategy introduces an additional hyperparameter, $threshold$, which can be tuned by looking at the average norm of the gradients during training \cite{pascanuDifficultyTrainingRecurrent2013}.

\section{Method} \label{sec:method}

In this section, we propose the Latent ODE-LSTM architecture and prove that the ODE-LSTM encoder mitigates the vanishing gradient problem. Moreover, we prove that the Latent ODE-LSTM suffers from the exploding gradient problem, and, to overcome this issue, we propose to incorporate the Norm Gradient Clipping strategy \cite{pascanuDifficultyTrainingRecurrent2013} into Latent ODE-LSTMs.

\subsection{Latent ODE-LSTM network}

We propose to replace the encoder in Latent ODE-RNNs with a ODE-LSTM \cite{lechnerLearningLongTermDependencies2020}, designated by Latent ODE-LSTMs. With the introduction of LSTMs into the encoder, we expect to mitigate the vanishing gradient problem and to improve performance in learning long-term dependencies.

The Latent ODE-LSTM is a VAE where the encoder combines an ODE-LSTM with a translator NN $g$ that approximates the posterior distribution $q(z_0|\{x_i,t_i\}_{i=1}^N)$ with mean $\mu$ and standard deviation $\sigma$:

\begin{equation*}
q_{\phi}(z_{0}| \{x_i,t_i\}_{i=1}^{N}) = \mathcal{N}(\mu,\sigma) \,\,\, \text{where} \,\,\ \mu,\sigma=g(z'_0) \,\,\ with \,\,\, z'_0 =\text{ODE-LSTM }_\phi(\{x_i,t_i\}_{i=1}^{N}).
\end{equation*}

Then the encoder converts the reversed input sequence $\{x_i,t_i\}_{i=1}^{N}$ into a latent variable $z_0$ at the initial time: 
\begin{equation*}
    z_0 = \mu + \sigma \odot \epsilon \in \mathbb{R}^l
\end{equation*}
\noindent where $l$ is the dimension of the latent space, $\phi$ are the parameters of the NN encoder and $\epsilon$ is randomly sampled noise.

The ODE-LSTM encoder is an LSTM where the transitions between two observations are given by a Neural ODE. At time step $i$, the LSTM cell is defined as follows:

\begin{equation}\label{eq:lstmODE}
\begin{array}{rcl}
   I_{i}&=  &   \sigma( \textbf{w}_{xin}x_i +  \textbf{w}_{hin}h'_i+b_{in})\\
   F_{i}&= &\sigma(  \textbf{w}_{xf}x_i+ \textbf{w}_{hf}h'_i +b_f) \\
    O_i&=&\sigma( \textbf{w}_{xo}x_i +  \textbf{w}_{ho}h'_i+b_o) \\

    \tilde{C}_{i}&=& \tanh(\textbf{w}_{xc} x_i  + \textbf{w}_{hc} h'_i +b_c)\\
    C_{i}&=& F_i \odot C_{i-1}+I_i \odot \tilde{C}_i  \\
    h_{i}&=& O_{i} \odot \tanh(C_i),
\end{array}
\end{equation}

\noindent where $h'_i \in \mathbb{R}^n$ is the intermediate hidden state, that is the solution of the ODE initial value problem with initial state $h_{i-1}$, and time interval $(t_{i-1}, t_i)$. In this case, the LSTM update formula \eqref{eq:lstmODE} can be represented by: 
$$(C_i,h_i) = LSTMCell(C_{i-1},h'_{i},x_i).$$

Algorithm \ref{alg:ODELSTM} describes the ODE-LSTM embed in the encoder of a Latent ODE-LSTM.

\begin{algorithm}[H]
\caption{The ODE-LSTM. The state transitions are continuous, unlike vanilla LSTM, and are the solution of an ODE initial value problem, computed by a Neural ODE.}
\label{alg:ODELSTM}
\begin{algorithmic}
\State \textbf{Input:} Data points and their timestamps $\{(x_i,t_i)\}_{i=1}^N$
\State $h_0=\textbf{0}$, $C_0=\textbf{0}$
\For{$i$ in 1,2,...,N }
\State $h'_i = ODESolve(f_\theta, h_{i-1}, (t_{i-1},t_i))$ 
\State $(C_i, h_i) = LSTMCell(C_{i-1}, h'_i,x_i)$
\EndFor
\State $\hat{y}_i = Output(h_i)$ for all $i=,1 \dots, N$
\State \textbf{Return:} $\{\hat{y}_i\}_{i=1}^N; h_N$
\end{algorithmic}
\end{algorithm}

Then, the latent state $z_0$ is given to the Neural ODE decoder that learns a continuous-time latent trajectory $f(z_0, \theta,(t_0,t_N))$ that allows prediction or extrapolation by solving an ODE initial value problem in the entire time domain ($t_0,\dots,t_N$):

\begin{equation*}
\{z_{t_i}\}_{i=0}^{N} = ODESolve(f_\theta, z_0, (t_{0}, \dots, t_N)).
\end{equation*}

Finally, the predicted latent states $\{z_{t_i}\}_{i=1..N}$ are fed into a neural network, which performs a conversion to the original data space and outputs the predicted data $\{\hat{y}_i\}_{i=1}^N$. The algorithm that describes the proposed Latent ODE-LSTM network is presented in Algorithm \ref{alg:latentODELSTM}.

\begin{algorithm}[H]
\caption{Latent ODE-LSTM. A VAE with an ODE-LSTM encoder and a Neural ODE decoder.}
\label{alg:latentODELSTM}
\begin{algorithmic}

\State \textbf{Input:} Data points and their timestamps $\{(x_i,t_i)\}_{i=1..N}$
\State $z'_0 = \text{ODE-LSTM} (\{x_i, t_i\}_{i=1..N})$
\State $\mu, \sigma = g(z'_0)$
\State $z_0 \sim \mathcal{N}(\mu, \sigma)$
\State $\{z_{t_i}\}_{i=1}^N = ODESolve(f_\theta,z_0,(t_0 \dots t_N))$
\State $\tilde{x}_i = OutputNN(z_{t_i})$ for all $i=1, \dots,N$
\State \textbf{Return:} $\{ \hat{y}_i \}_{i=1}^N$

\end{algorithmic}
\end{algorithm}

The training of Latent ODE-LSTMs is done by optimising the parameters of the encoder, $\phi$, and decoder, $\theta$, simultaneously. This is done by optimising the cost function defined in \eqref{eq:loss}.

\subsection{Latent ODE-LSTMs prevent the vanishing gradient problem} \label{proof:Latent ODE-LSTM vanishing}

To find the parameters $\Theta$ that optimise the loss $\mathcal{L}_\Theta$, the gradients with respect to the parameters are computed as follows:

\begin{equation}
\label{eq:ODELSTM1.0}
    \dfrac{\partial \mathcal{L} }{\partial \Theta}= \frac{1}{N} \sum_{i=1}^{N} \dfrac{\partial l(\hat{y}_i)}{\partial \Theta}
\end{equation}

\noindent where $\dfrac{\partial l(\hat{y}_i)}{\partial \Theta}$ is a sum of products that gives the gradient at time step $i$, which is given by all the contributions of the previous time steps $k$, with $k<i$:
\begin{equation}
\label{eq:ODELSTM1.1}
\dfrac{\partial l(\hat{y}_i)}{\partial \Theta} = \sum_{k=1}^i  \dfrac{\partial l(\hat{y}_i)}{\partial \hat{y}_i} \dfrac{\partial \hat{y}_i}{\partial \text{O}_{\text{NN}}} \dfrac{\partial \text{O}_{\text{NN}}}{\partial z_{t_i}} \dfrac{\partial z_{t_i}}{\partial z_0} \left(\dfrac{\partial z_0}{\partial \mu} \dfrac{\partial \mu}{\partial g_\text{NN}} \dfrac{\partial g_\text{NN}}{\partial z'_0} \dfrac{\partial z'_0}{\partial h_i} + \dfrac{\partial z_0}{\partial \sigma} \dfrac{\partial \sigma}{\partial g_\text{NN}} \dfrac{\partial g_\text{NN}}{\partial z'_0} \dfrac{\partial z'_0}{\partial h_i}  \right) \dfrac{\partial h_i }{\partial h_k} \dfrac{\partial^+ h_k}{\partial \Theta},
\end{equation}

\noindent where $\dfrac{\partial^+ h_k}{\partial \theta}$ is the \textit{immediate} partial derivative of $h_k$ with respect to the parameters $\Theta$ \cite{pascanuDifficultyTrainingRecurrent2013}.

Here,  $\dfrac{\partial h_i}{\partial h_{k}}$ is a chain of products of all the hidden states that contribute to the hidden state $h_i$, at time step $i$:

\begin{equation}
\label{eq:ODELSTM2}
    \frac{\partial h_i}{\partial h_k} = \left( \prod_{i \geq j>k} \frac{\partial h_j}{\partial h'_{j}} \frac{\partial h'_{j}}{\partial h_{j-1}} \right).
\end{equation}

 In Latent ODE-LSTMs, the ODE-RNN is replaced by an ODE-LSTM in the encoder. In this case the update of the LSTM cell \eqref{eq:lstmODE} are more complex, where the hidden state at time step $j$ is given by:

\begin{equation}
\label{eq:0}    
h_{j} = O_{j} \odot \tanh(\underbrace{F_{j} \odot C_{j-1}+I_{j} \odot \tilde{C}_{j}}_{C_j}).
\end{equation}

The term $\dfrac{\partial h_j}{\partial h'_{j}}$  is given by a chain of products (see \ref{eq:ODELSTM2}). Considering \eqref{eq:0} to compute the $\dfrac{\partial h_j}{\partial h'_j}$, the vectors are converted into diagonal matrices and the Hadamard product is replaced by the usual product,  the derivatives are computed in an element-wise fashion:

\begin{equation}
\label{eq:3}
    \dfrac{\partial h_j}{\partial h'_{j}} = \dfrac{\partial}{\partial h'_{j}} \left( O_j \odot \tanh(C_j) \right) = \dfrac{\partial}{\partial h'_{j}} \left( \textbf{O}_j  diag(\tanh(C_j)) \right) = \dfrac{\partial \textbf{O}_j}{\partial h'_{j}}  diag(\tanh(C_j)) + \textbf{O}_j  diag \left(\dfrac{\partial \tanh(C_j)}{\partial h'_{j}} \right)
\end{equation}

\noindent where $\textbf{O}_j=diag(O_j),$
\begin{equation}
\label{eq:4}
    \dfrac{\partial \textbf{O}_j}{\partial h'_{j}} = 
    \textbf{w}_{ho}  diag(\sigma'( \textbf{w}_{xo} x_j +  \textbf{w}_{ho} h'_{j} + b_o))
\end{equation}

\noindent and 
\begin{equation}
\label{eq:5}
\begin{array}{rcl}
    
    diag \left( \dfrac{\partial \tanh(C_j)}{\partial h'_{j}}\right)& =& diag\left(\dfrac{\partial}{\partial h'_{j}} \left( \tanh(F_j \odot C_{j-1} + I_j \odot \tilde{C}_j) \right) \right)\\ 
    & =& diag\left(\dfrac{\partial}{\partial h'_{j}} \left( \tanh(\textbf{F}_j  \textbf{C}_{j-1} + \textbf{I}_j  \tilde{\textbf{C}}_j) \right) \right) \\ 
    &=& diag \left(\tanh'(\textbf{F}_j  \textbf{C}_{j-1} + \textbf{I}_j  \tilde{\textbf{C}}_j)\right)  \left(\dfrac{\partial \textbf{F}_j}{\partial h'_{j}}  \textbf{C}_{j-1} + \textbf{F}_j  \dfrac{\partial \textbf{C}_{j-1}}{\partial h'_{j}} + \dfrac{\partial \textbf{I}_j}{\partial h'_{j}}  \tilde{\textbf{C}}_j + \textbf{I}_j  \dfrac{\partial \tilde{\textbf{C}}_j}{\partial h'_{j}} \right)
\end{array}
\end{equation}
with  $\textbf{F}_j=diag(F_j), \textbf{C}_{j-1}=diag(C_{j-1}), \textbf{I}_j=diag(I_j)$  $\tilde{\textbf{C}}_j=diag(\tilde{C}_j)$, and where

\begin{equation}
\label{eq:6}
    \dfrac{\partial \textbf{F}_j}{\partial h'_{j}} = \textbf{w}_{hf}  diag\left(\sigma'(  \textbf{w}_{xf} x_j+  \textbf{w}_{hf} h'_{j} + b_f)\right),
\end{equation}

\begin{equation}
\label{eq:7}
    \dfrac{\partial \textbf{I}_j}{\partial h'_{j}} = \textbf{w}_{h\text{in}}  diag\left(\sigma'( \textbf{w}_{x\text{in}} x_j +  \textbf{w}_{h\text{in}} h'_{j} + b_{\text{in}})\right),
\end{equation}

\begin{equation}
\label{eq:8}
    \dfrac{\partial \tilde{\textbf{C}}_j}{\partial h'_{j}} = \textbf{w}_{hc}  diag\left(\tanh'( \textbf{w}_{xc} x_j +  \textbf{w}_{hc} h'_{j} + b_{c})\right).
\end{equation}

Substituting \eqref{eq:6}-\eqref{eq:8} into \eqref{eq:5}, we obtain:

\begin{equation}
\label{eq:10}
\begin{array}{rcl}
 diag \left( \dfrac{\partial \tanh(C_j)}{\partial h'_{j}} \right) &=& diag\left(\tanh'(\textbf{F}_j  \textbf{C}_{j-1} + \textbf{I}_j  \tilde{\textbf{C}}_j)\right)  [  \textbf{w}_{hf}  diag(\sigma'(  \textbf{w}_{xf} x_j+ \textbf{w}_{hf} h'_{j}  + b_f))  \textbf{C}_{j-1} +\\
 & &  \textbf{F}_j  \dfrac{\partial \textbf{C}_{j-1}}{\partial h'_{j}} + \textbf{w}_{h\text{in}}  diag(\sigma'(  \textbf{w}_{x\text{in}} x_j+  \textbf{w}_{h\text{in}} h'_{j} + b_{\text{in}}))  \tilde{\textbf{C}}_j+\\
 & & \textbf{I}_j  \textbf{w}_{hc}  diag(\tanh'( \textbf{w}_{xc} x_j +  \textbf{w}_{hc} h'_{j} + b_{c})) ].    
\end{array}
\end{equation}

Substituting \eqref{eq:4} and \eqref{eq:10} into \eqref{eq:3}, we obtain $\dfrac{\partial h_j}{\partial h'_j}$:

\begin{equation}
\begin{array}{rcl}
\label{eq:11}
    \dfrac{\partial h_j}{\partial h'_{j}} & = &\textbf{w}_{ho}  diag(\sigma'(\textbf{w}_{xo} x_j  +  \textbf{w}_{ho} h'_{j} + b_o))  diag(\tanh(C_j)) + \textbf{O}_j  [diag(\tanh'(\textbf{F}_j  \textbf{C}_{j-1} + \textbf{I}_j  \tilde{\textbf{C}}_j)) \times \\ 
    & &[  \textbf{w}_{hf}  diag(\sigma'(\textbf{w}_{xf} x_j  +  \textbf{w}_{hf} h'_{j} + b_f))  \textbf{C}_{j-1} + \textbf{F}_j  \dfrac{\partial \textbf{C}_{j-1}}{\partial h'_{j}} + \\
    & & \textbf{w}_{h\text{in}}  diag(\sigma'(\textbf{w}_{x\text{in}} x_j  +  \textbf{w}_{h\text{in}} h'_{j} + b_{\text{in}})) + 
     \tilde{\textbf{C}}_j + \textbf{I}_j  \textbf{w}_{hc}  diag(\tanh'(\textbf{w}_{xc} x_j  + \textbf{w}_{hc} h'_{j}  + b_{c})) ]].    
\end{array}
\end{equation}

 Substituting \eqref{eq:11} into \eqref{eq:ODELSTM2}, and converting the $\dfrac{\partial h'_j}{\partial h_{j-1}}$ into a diagonal matrix, we obtain:

 \begin{equation}
\label{eq:11a}
     \dfrac{\partial h_i}{\partial h_k} =  \prod_{i \geq j \geq k} \dfrac{\partial h_j}{\partial h'_j}\, diag\left(\dfrac{\partial h'_j}{\partial h_{j-1}}\right) 
 \end{equation}
where $\dfrac{\partial h_j}{\partial h'_j}$ is given by \eqref{eq:11}.

Note that, the behaviour of the sum in \eqref{eq:ODELSTM1.0} is given by the behaviour of the terms $\dfrac{\partial l(\hat{y}_i)}{\partial \Theta}$, which have all the same form. Each temporal contribution is given by \eqref{eq:ODELSTM1.1} and measures how parameters $\Theta$ at time step $k$ affect the loss at time step $i>k$. The factors $\dfrac{h_i}{h_k}$ \eqref{eq:ODELSTM2} transport the error \emph{in time} from step $i$ back to step $k$. To analyse the vanishing and exploding gradient problems, we need to look in particular to these matrix factors that takes the form of a product of $i-k$ Jacobian matrices.

Although less prone to this problem, Latent ODE-LSTM can still suffer from the vanishing gradient problem:

\begin{multline}
\label{eq:12}
\forall j, \left \lVert \dfrac{\partial h_j}{\partial h'_j}\, diag\left(\dfrac{\partial h'_j}{\partial h_{j-1}}\right) \right \rVert=\left \lVert \dfrac{\partial \textbf{O}_j}{\partial h'_{j}}  diag(\tanh(C_j)) + \textbf{O}_j  diag \left( \dfrac{\partial \tanh(C_j)}{\partial h'_{j}} \right) \right \rVert \left \lVert diag \left( \dfrac{\partial h'_j}{\partial h_{j-1}} \right) \right \rVert \leq \\ \left \lVert \dfrac{\partial \textbf{O}_j}{\partial h'_{j}}  diag(\tanh(C_j))   + \textbf{O}_j  diag \left(\dfrac{\partial \tanh(C_j)}{\partial h'_{j}}\right)  \right \rVert \left \lVert diag \left( \dfrac{\partial h'_j}{\partial h_{j-1}} \right) \right \rVert.
\end{multline}

Let $\lambda_{ho}, \lambda_{hf}, \lambda_{h\text{in}}, \lambda_{hc}, \lambda, \lambda_d, \lambda'$ be the absolute value of the largest eigenvalue of the weight matrix of the output $\textbf{w}_{ho}$, forget $\textbf{w}_{hf}$ and input $\textbf{w}_{h\text{in}}$ gates and candidate cell $\textbf{w}_{hc}$, and of the matrix $\textbf{C}_{j-1}$, $\dfrac{\partial \textbf{C}_{j-1}}{\partial h'_j}$ and $\dfrac{\partial h'_j}{\partial h_{j-1}}$,  respectively. 

Let $\gamma_{t}$  and $\gamma_{\sigma}$ be the absolute values of $\tanh(.)$ and $\sigma(.)$ and therefore $\lVert diag(\tanh(.)\rVert \leq \gamma_t$,  $\lVert diag(\sigma(.)\rVert \leq \gamma_\sigma$, respectively.

Let $\gamma_{td}$  and $\gamma_{\sigma d}$ be the absolute values of $\tanh'(.)$ and $\sigma'(.)$ and therefore $\lVert diag(\tanh'(.)\rVert \leq \gamma_{td}$,  $\lVert diag(\sigma'(.)\rVert \leq \gamma_{\sigma d}$, respectively.

For the gradient to vanish, the chain of products given by \eqref{eq:11a} would have to decrease linearly toward zero \cite{hochreiterLongShorttermMemory1997}. For this to happen, it suffices that

\begin{equation}\nonumber
    \begin{array}{l}
        \left [ \lambda_{ho} \gamma_{\sigma d} \gamma_t + \gamma_\sigma ( \gamma_{td} \lambda_{hf} \gamma_{\sigma d} \lambda + \gamma_\sigma \lambda_d + \lambda_{h\text{in}} \gamma_{\sigma d} \gamma_t + \gamma_{\sigma d} \lambda_{hc} \gamma_{td} ) \right ] \lambda'<\\
         \hspace{1cm} <\left [ 1 + \gamma_\sigma ( 1 + 1 + 1 + 1 ) \right ] \lambda'\\
         \hspace{1cm} <\left [ 1 + 4\gamma_\sigma  \right ] \lambda'\\
         \hspace{1cm} < 5 \lambda'\\
         \hspace{1cm} < 1
    \end{array}
\end{equation}

with $\lambda_{ho} < \dfrac{1}{\gamma_{\sigma d} \gamma_t}, \lambda_{hf} \lambda < \dfrac{1}{\gamma_{td} \gamma_{\sigma d}}, \lambda_d < \dfrac{1}{\gamma_\sigma}, \lambda_{h\text{in}} < \dfrac{1}{\gamma_{\sigma d} \gamma_t}, \lambda_{hc} < \dfrac{1}{\gamma_{\sigma d} \gamma_{td}}$ and $\lambda' < \dfrac{1}{5}.$

For all $j$, taking $\eta \in \mathbb{R}$ so that $\eta < 1$ comes $\left \lVert  \dfrac{\partial h_j}{\partial h'_j} diag \left(\dfrac{\partial h'_j}{\partial h_{j-1}}\right) \right \rVert \leq \eta < 1$. By induction over $j$ we obtain

\begin{multline}
\label{eq:12.1}
    \dfrac{\partial l(\hat{y}_i)}{\partial \hat{y}_i} \dfrac{\partial \hat{y}_i}{\partial \text{O}_{\text{NN}}} \dfrac{\partial \text{O}_{\text{NN}}}{\partial z_{t_i}} \dfrac{\partial z_{t_i}}{\partial z_0} \left(\dfrac{\partial z_0}{\partial \mu} \dfrac{\partial \mu}{\partial g_\text{NN}} \dfrac{\partial g_\text{NN}}{\partial z'_0} \dfrac{\partial z'_0}{\partial h_i} + \dfrac{\partial z_0}{\partial \sigma} \dfrac{\partial \sigma}{\partial g_\text{NN}} \dfrac{\partial g_\text{NN}}{\partial z'_0} \dfrac{\partial z'_0}{\partial h_i}  \right) \left( \prod_{i \geq j>k} \frac{\partial h_j}{\partial h'_{j}} \frac{\partial h'_{j}}{\partial h_{j-1}} \right) \leq \\
    \leq \dfrac{\partial l(\hat{y}_i)}{\partial \hat{y}_i} \dfrac{\partial \hat{y}_i}{\partial \text{O}_{\text{NN}}} \dfrac{\partial \text{O}_{\text{NN}}}{\partial z_{t_i}} \dfrac{\partial z_{t_i}}{\partial z_0} \left(\dfrac{\partial z_0}{\partial \mu} \dfrac{\partial \mu}{\partial g_\text{NN}} \dfrac{\partial g_\text{NN}}{\partial z'_0} \dfrac{\partial z'_0}{\partial h_i} + \dfrac{\partial z_0}{\partial \sigma} \dfrac{\partial \sigma}{\partial g_\text{NN}} \dfrac{\partial g_\text{NN}}{\partial z'_0} \dfrac{\partial z'_0}{\partial h_i}  \right) {\eta}^{i-k} .
\end{multline}

As $\eta < 1$ it follows from \eqref{eq:12.1} that in the presence of long-term sequences of data (for which $i-k$ is large) then the term ${\eta}^{i-k}$ decreases exponentially to zero, leading to the vanishing of the gradients.

Note that comparing the term $\dfrac{\partial h_i}{\partial h_k}$ \eqref{eq:11a} with the same term in \eqref{eq:odernn5} in \ref{proof:ODE-RNN}, highlights the difference between the computation of a Latent ODE-LSTM and a Latent ODE-RNN networks. 

Although it has the same structure, $\dfrac{\partial h_i}{\partial h_k}$ is no longer a simple chain of products, but a chain of products of sums \eqref{eq:11a}. This ensures constant error flow through the network due to the dependence of the previous memory cell $C_{j-1}$ with the current $C_j$, preventing gradients from vanishing \cite{hochreiterLongShorttermMemory1997}.

\subsection{Latent ODE-LSTMs suffer from the exploding gradient problem} \label{proof:Latent ODE-LSTM exploding}

LSTMs cannot eliminate or mitigate the exploding gradient problem \cite{pascanuDifficultyTrainingRecurrent2013, sutskeverSequenceSequenceLearning2014} and the problem propagates to Latent ODE-LSTMs due to the LSTM update scheme used by the ODE-LSTM encoder.

Consider the absolute value of the largest eigenvalues defined in the vanishing gradient proof shown previously. 
It is necessary that 

\begin{equation}\nonumber
    \begin{array}{l}
        \left [ \lambda_{ho} \gamma_{\sigma d} \gamma_t + \gamma_\sigma ( \gamma_{td} \lambda_{hf} \gamma_{\sigma d} \lambda + \gamma_\sigma \lambda_d + \lambda_{h\text{in}} \gamma_{\sigma d} \gamma_t + \gamma_{\sigma d} \lambda_{hc} \gamma_{td} ) \right ] \lambda'>\\
         \hspace{1cm} >\left [ 1 + \gamma_\sigma ( 1 + 1 + 1 + 1 ) \right ] \lambda'\\
         \hspace{1cm} >\left [ 1 + 4\gamma_\sigma  \right ] \lambda'\\
         \hspace{1cm} > 5 \lambda'\\
         \hspace{1cm} > 1
    \end{array}
\end{equation}

with $\lambda_{ho} > \dfrac{1}{\gamma_{\sigma d} \gamma_t}, \lambda_{hf} \lambda > \dfrac{1}{\gamma_{td} \gamma_{\sigma d}}, \lambda_d > \dfrac{1}{\gamma_\sigma}, \lambda_{h\text{in}} > \dfrac{1}{\gamma_{\sigma d} \gamma_t}, \lambda_{hc} > \dfrac{1}{\gamma_{\sigma d} \gamma_{td}}$ and $\lambda' > \dfrac{1}{5}.$

Let $\eta \in \mathbb{R}$ so that $\eta>1$, comes $\forall j, \left \lVert  \dfrac{\partial h_j}{\partial h'_j}\, diag\left(\dfrac{\partial h'_j}{\partial h_{j-1}}\right)  \right \rVert \geq \eta > 1$. By induction over $j$, we obtain 

\begin{multline}
\label{eq:13}
    \dfrac{\partial l(\hat{y}_i)}{\partial \hat{y}_i} \dfrac{\partial \hat{y}_i}{\partial \text{O}_{\text{NN}}} \dfrac{\partial \text{O}_{\text{NN}}}{\partial z_{t_i}} \dfrac{\partial z_{t_i}}{\partial z_0} \left(\dfrac{\partial z_0}{\partial \mu} \dfrac{\partial \mu}{\partial g_\text{NN}} \dfrac{\partial g_\text{NN}}{\partial z'_0} \dfrac{\partial z'_0}{\partial h_i} + \dfrac{\partial z_0}{\partial \sigma} \dfrac{\partial \sigma}{\partial g_\text{NN}} \dfrac{\partial g_\text{NN}}{\partial z'_0} \dfrac{\partial z'_0}{\partial h_i}  \right) \left( \prod_{i \geq j>k} \frac{\partial h_j}{\partial h'_{j}} \frac{\partial h'_{j}}{\partial h_{j-1}} \right) \geq \\
    \geq  \dfrac{\partial l(\hat{y}_i)}{\partial \hat{y}_i} \dfrac{\partial \hat{y}_i}{\partial \text{O}_{\text{NN}}} \dfrac{\partial \text{O}_{\text{NN}}}{\partial z_{t_i}} \dfrac{\partial z_{t_i}}{\partial z_0} \left(\dfrac{\partial z_0}{\partial \mu} \dfrac{\partial \mu}{\partial g_\text{NN}} \dfrac{\partial g_\text{NN}}{\partial z'_0} \dfrac{\partial z'_0}{\partial h_i} + \dfrac{\partial z_0}{\partial \sigma} \dfrac{\partial \sigma}{\partial g_\text{NN}} \dfrac{\partial g_\text{NN}}{\partial z'_0} \dfrac{\partial z'_0}{\partial h_i}  \right) {\eta}^{i-k}.
\end{multline}

As $\eta > 1$, it follows from \eqref{eq:13} that in the presence of long-term sequences of data (for which $i-k$ is large), the term ${\eta}^{i-k}$ increases exponentially with $i-k$, leading to the explosion of the gradients. 
\\

\subsection{Latent ODE-LSTM + Gradient clipping}

As shown in Section \ref{proof:Latent ODE-LSTM exploding}, Latent ODE-LSTMs suffer from the exploding gradient problem. To prevent this from happening, we propose to augment the Latent ODE-LSTM architecture with the Norm Gradient Clipping \cite{pascanuDifficultyTrainingRecurrent2013}. This strategy can be easily implemented in the training loop used in most Deep Learning libraries and does not require any change in the implementation of the Latent ODE-LSTM network.

\section{Experiments} \label{sec:experiments}

To analyse the performance of the Latent ODE-LSTM model, two different evaluations were carried out: a qualitative approach to visually test time series reconstruction and extrapolation using $4$ datasets with irregularly sampled time points forming spirals; and a quantitative evaluation using Mean Squared Error (MSE) in extrapolation tasks. Two datasets consisting of real-life time series data publicly available at Kaggle \cite{kaggle} were used. Both datasets are sampled daily, one of them regularly, while the other is sampled irregularly.

To evaluate the performance of the newly proposed architecture with and without gradient clipping, a Latent ODE-RNN was used as a baseline.

The Latent ODE-LSTM (with and without gradient clipping) models and Latent ODE-RNN model were implemented from scratch in \textit{Pytorch}, available at \href{https://github.com/CeciliaCoelho/LatentODELSTM}{github.com/CeciliaCoelho/LatentODELSTM}.

\subsection{Qualitative evaluation}

A qualitative assessment was also carried out in \cite{rubanovaLatentOrdinaryDifferential2019} to test the performance of Latent ODE and Latent ODE-RNN. Here, we compare the performance of our Latent ODE-LSTM and Latent ODE-LSTM+GC with Latent ODE-RNN \cite{rubanovaLatentOrdinaryDifferential2019} on the spirals dataset.

\subsubsection{Experimental Conditions}

The VAE models for the qualitative evaluation consist of: ODE-LSTM (or ODE-RNN) encoder with $1$ hidden layer and $20$ neurons, a Neural ODE with $1$ hidden layer and $20$ neurons and an output layer with $25$ neurons; and a Neural ODE decoder with $1$ hidden layer and $20$ neurons and an output network with $1$ hidden layer and $20$ neurons.
We chose Adam Optimiser with learning rate of $0.01$, batch size of $1000$ and $750$ epochs.
The Neural ODEs are solved with Fourth-order Runge-Kutta method with 3/8 rule and using a fixed-step. A fixed-step was preferred over an adaptive one since the ODE is stiff and the step size could go to zero.

\subsubsection{Bidirectional spiral dataset}

The bidirectional spiral dataset was introduced in \cite{chenNeuralOrdinaryDifferential2019} to test the performance of Latent ODEs in reconstructing and extrapolating spirals with irregularly sampled (forward and backward) time points.

\begin{figure}[H]
    \centering
    \resizebox{0.4\textwidth}{!}{
    \includegraphics{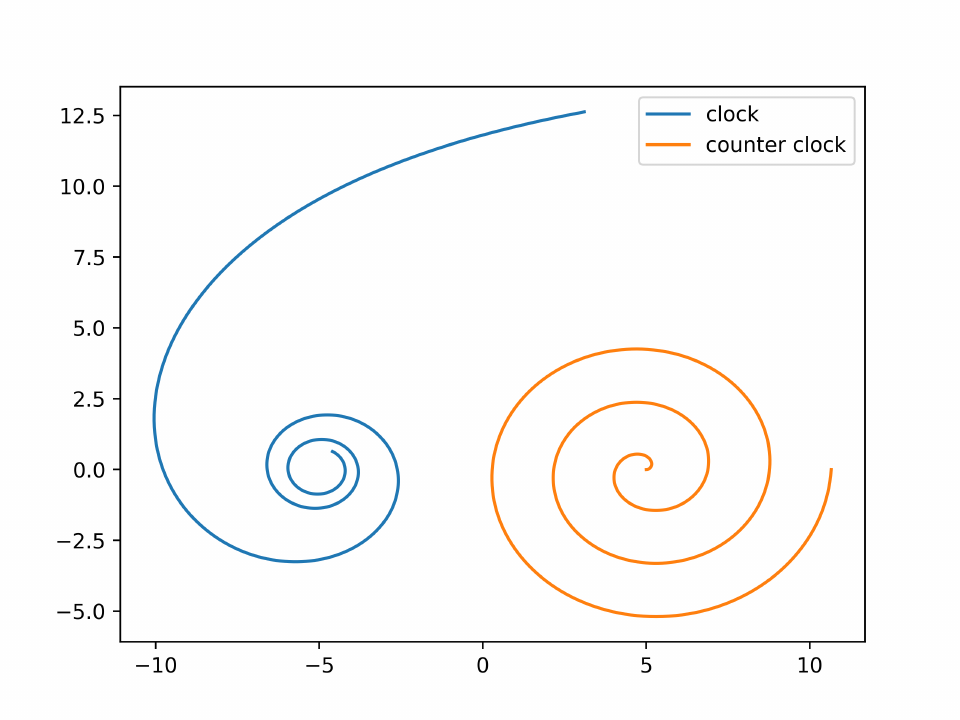}
    }
    \caption{Clockwise and counterclockwise spirals used for dataset generation.}
    \label{fig:spiralTruth}
\end{figure}

In this work, based on the code available in \cite{chenNeuralOrdinaryDifferential2019}, 
first a dataset with $500$ sequences taken from the clockwise 2D spirals and $500$ from the counterclockwise 2D spirals, with different starting points were generated. Each sequence has $500$ time points regularly sampled (Figure \ref{fig:spiralTruth} shows the clockwise and counterclockwise spiral), denoted by $D_{N=500}$. Then, from $D_{N=500}$ we constructed $4$ training datasets, each with different sequence length $N=\{30,50,100,250\}$ of randomly selected time points to which Gaussian noise was added, denoted by $D_{N=30},D_{N=50},D_{N=100},D_{N=250}$. Each training dataset has a total of $1000$ sequences.

We note that during the training of models Latent ODE-RNN and Latent ODE-LSTM without the gradient clipping strategy, when using $D_{N=250}$, the exploding gradient problem occurred. To overcome this, several trials of initialisation of the weights parameters were tested. For the Latent ODE-LSTM+GC, as expected, this issue did not occur. 

Reducing $N$ allows us to test the performance of the three models on progressively sparser data. 

\begin{figure}[H]
    \begin{subfigure}[b]{.33\textwidth}
    \includegraphics[width=\textwidth]{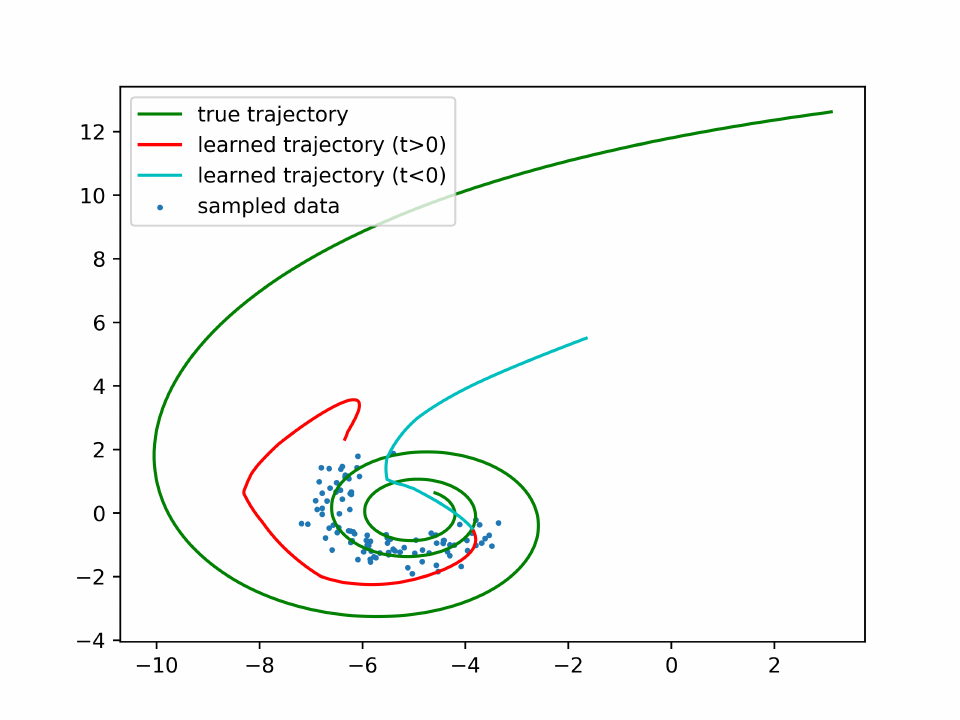} \\ \includegraphics[width=\textwidth]{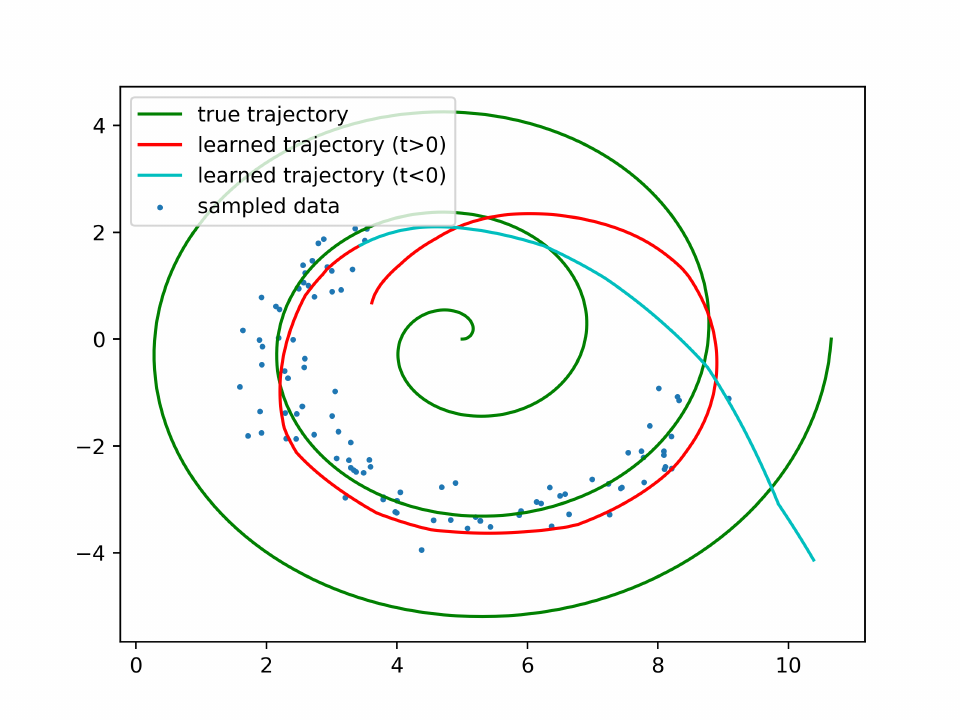}
    \caption{Latent ODE-RNN}
    \label{fig:spiral250a}
    \end{subfigure}
    \begin{subfigure}[b]{.33\textwidth}
    \includegraphics[width=\textwidth]{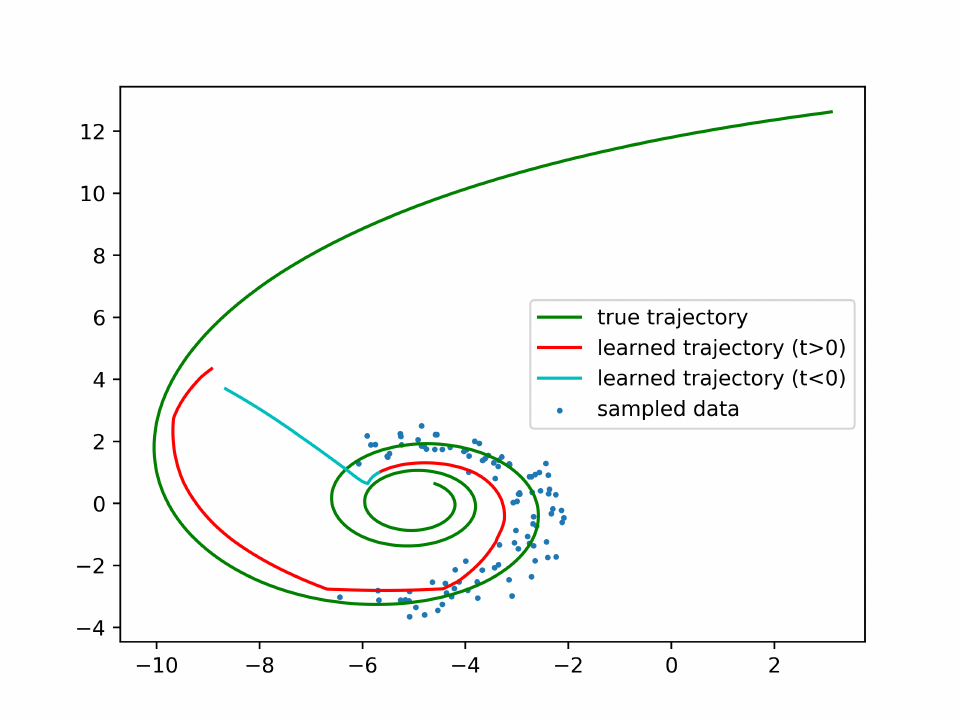} \\ \includegraphics[width=\textwidth]{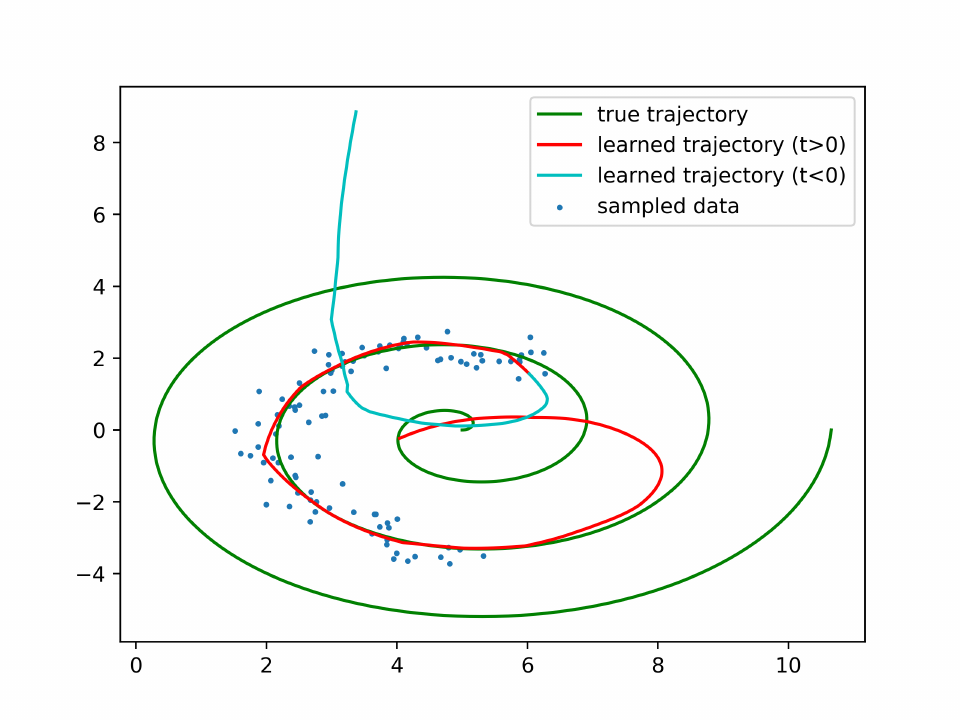} 
    \caption{Latent ODE-LSTM}
    \label{fig:spiral250b}
    \end{subfigure}
    \begin{subfigure}[b]{.33\textwidth}
    \includegraphics[width=\textwidth]{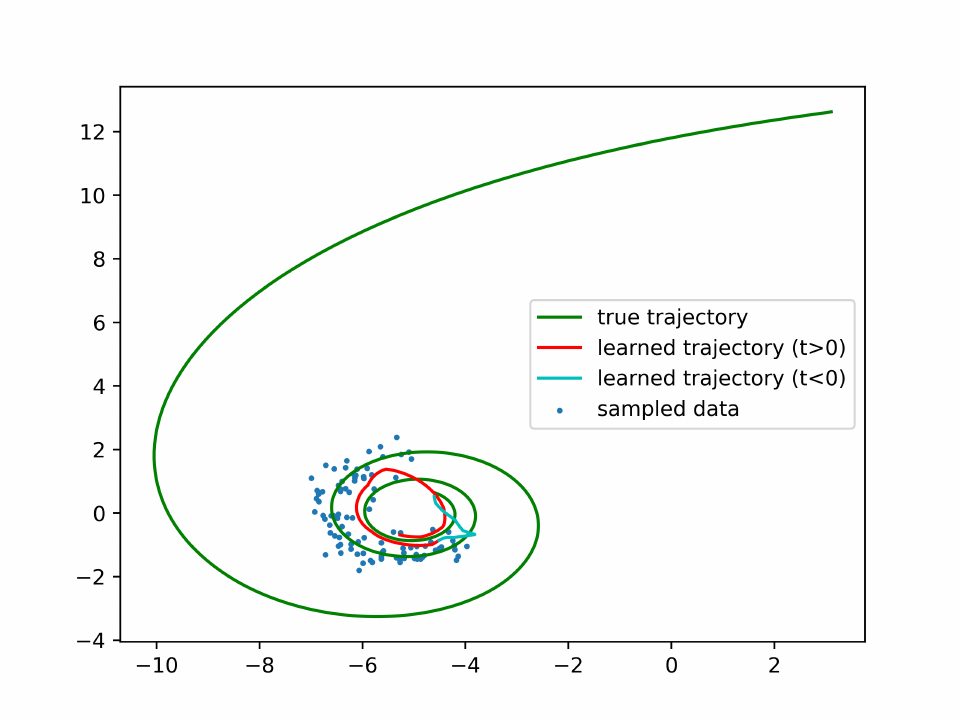} \\ \includegraphics[width=\textwidth]{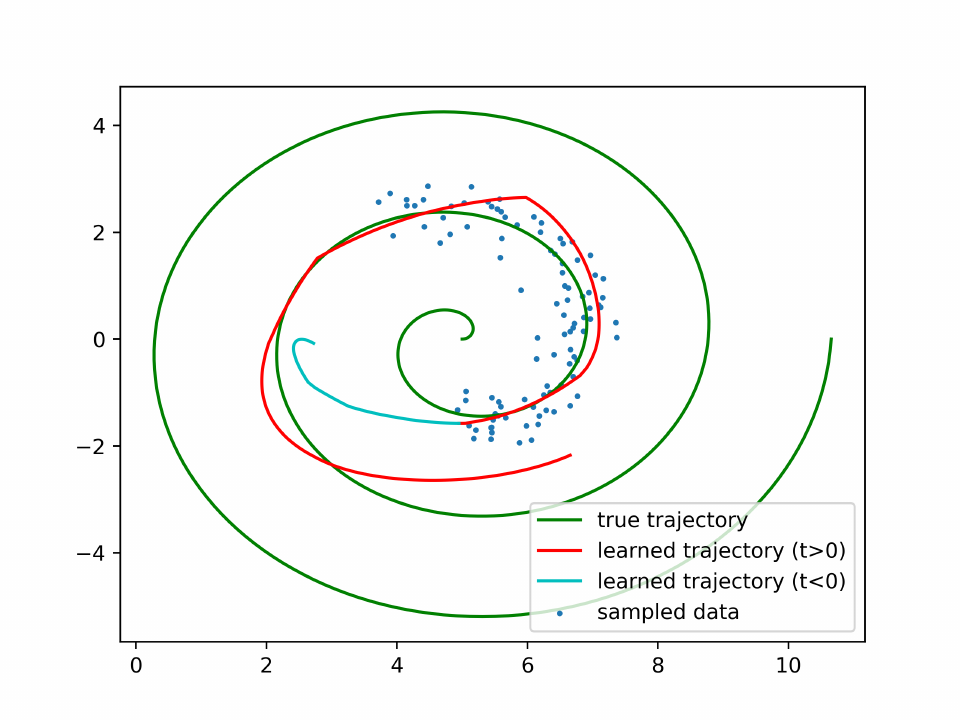} 
    \caption{Latent ODE-LSTM+GC}
    \label{fig:spiral250c}
    \end{subfigure}
    \caption{Visual performance of Latent ODE-RNN (a), Latent ODE-LSTM (b) and Latent ODE-LSTM+GC-Gradient Clipping (c) in the task of reconstruction (best fit to the true trajectory for the sampled data time points) and extrapolation (backwards in time, $t< 0$, shown in blue, and forward in time, $t > 0$, in red) of counterclockwise (top) and clockwise (bottom) spirals, after training with datasets with $N=250$.}
    \label{fig:spiral250}
\end{figure}

\begin{figure}[H]
    \begin{subfigure}[b]{.33\textwidth}
    \includegraphics[width=\textwidth]{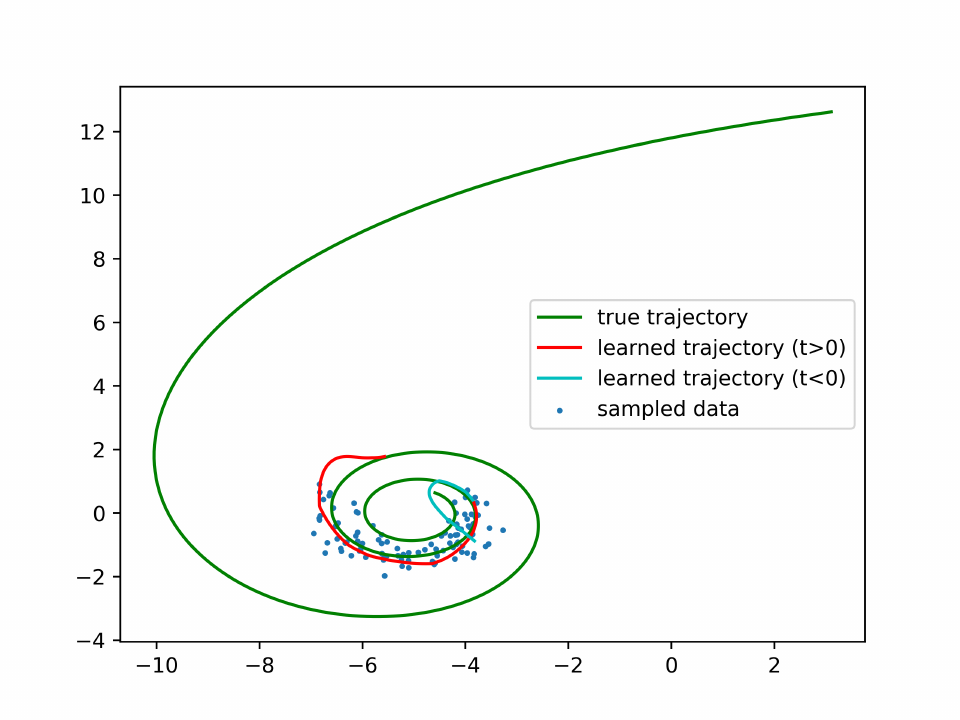} \\ \includegraphics[width=\textwidth]{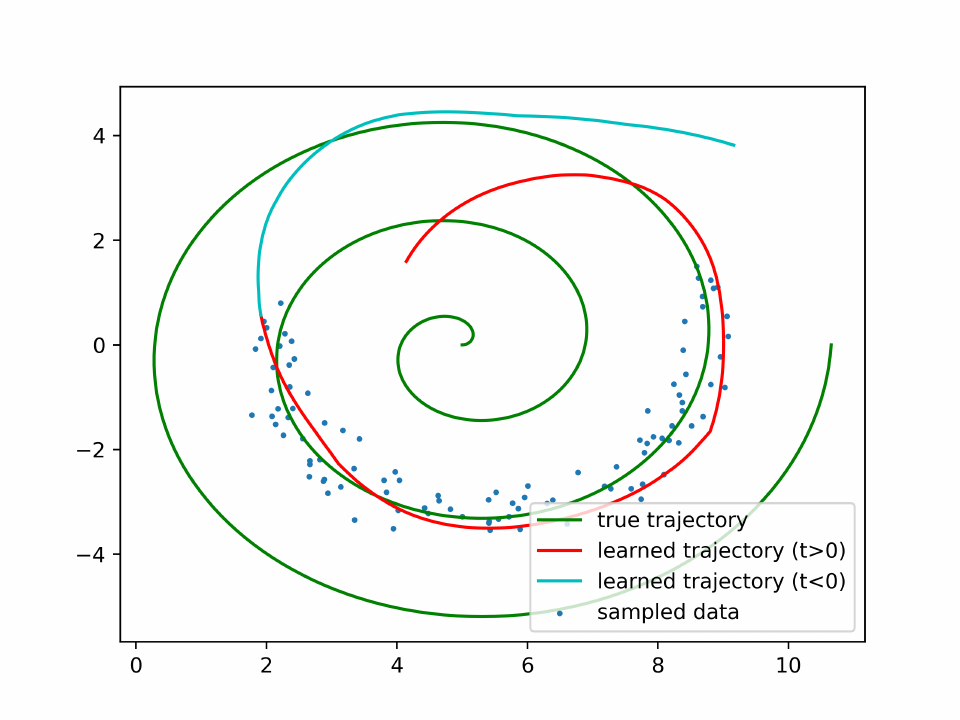}
    \caption{Latent ODE-RNN}
    \label{fig:spiral100a}
    \end{subfigure}
    \begin{subfigure}[b]{.33\textwidth}
    \includegraphics[width=\textwidth]{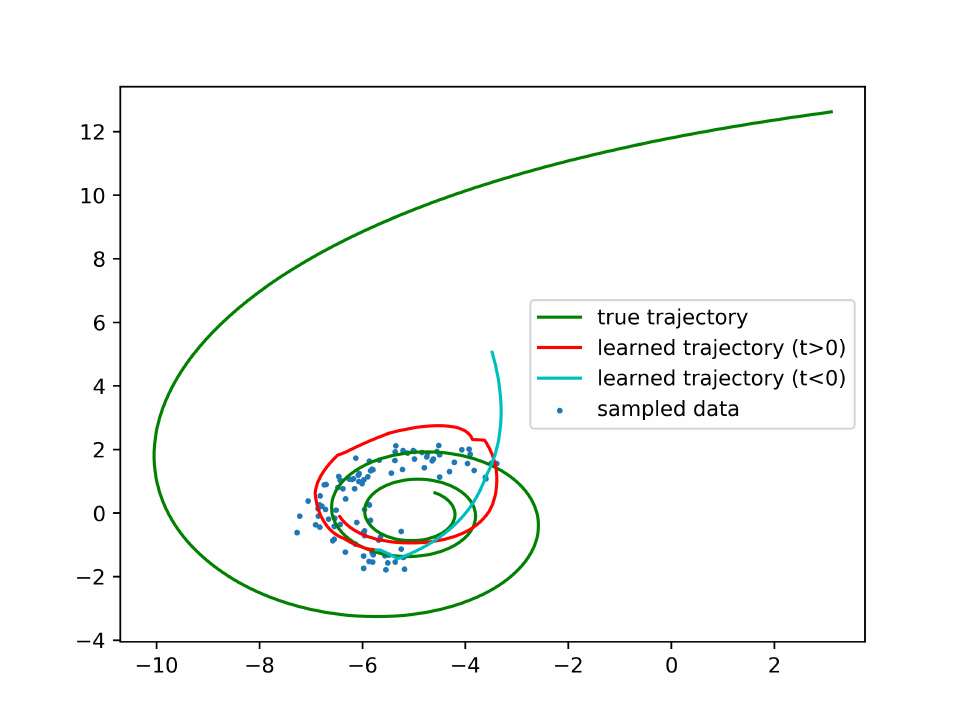} \\ \includegraphics[width=\textwidth]{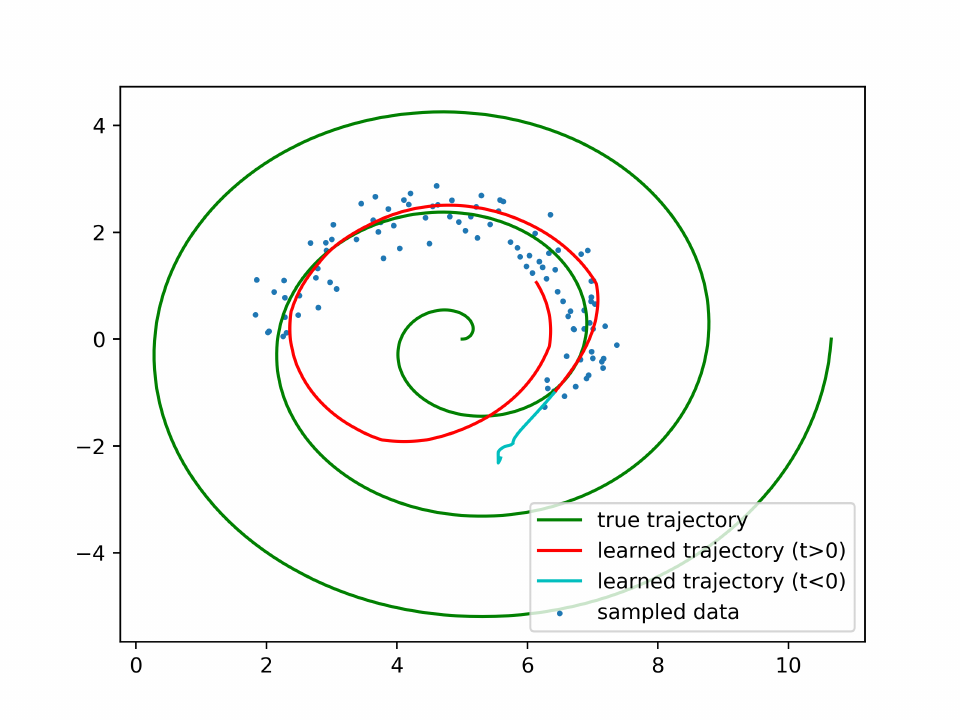}
    \caption{Latent ODE-LSTM}
    \label{fig:spiral100b}
    \end{subfigure}
    \begin{subfigure}[b]{.33\textwidth}
    \includegraphics[width=\textwidth]{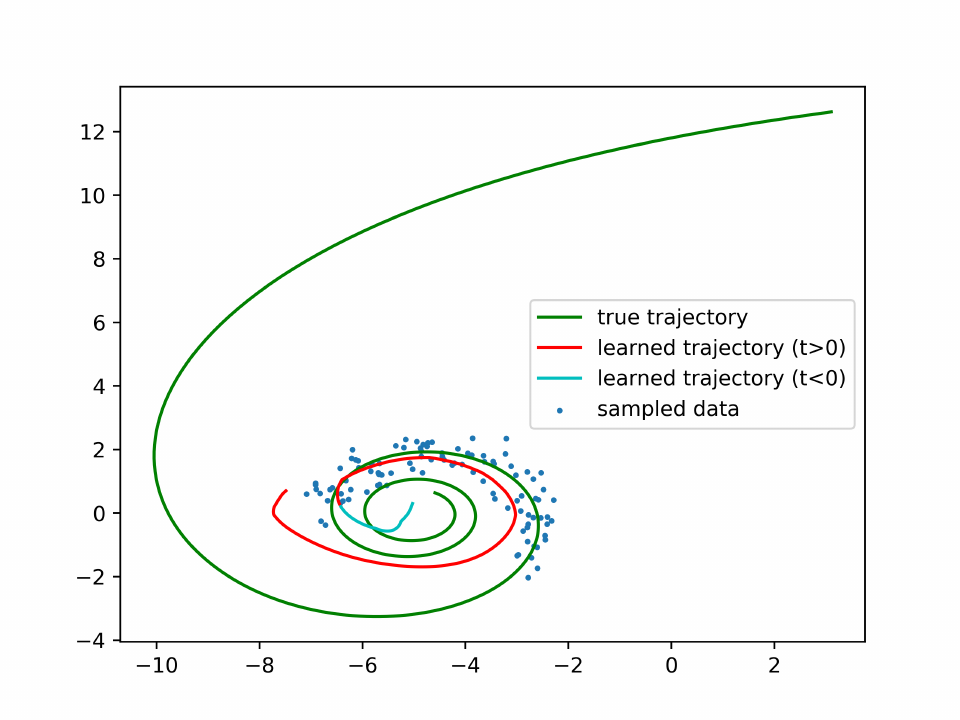} \\ \includegraphics[width=\textwidth]{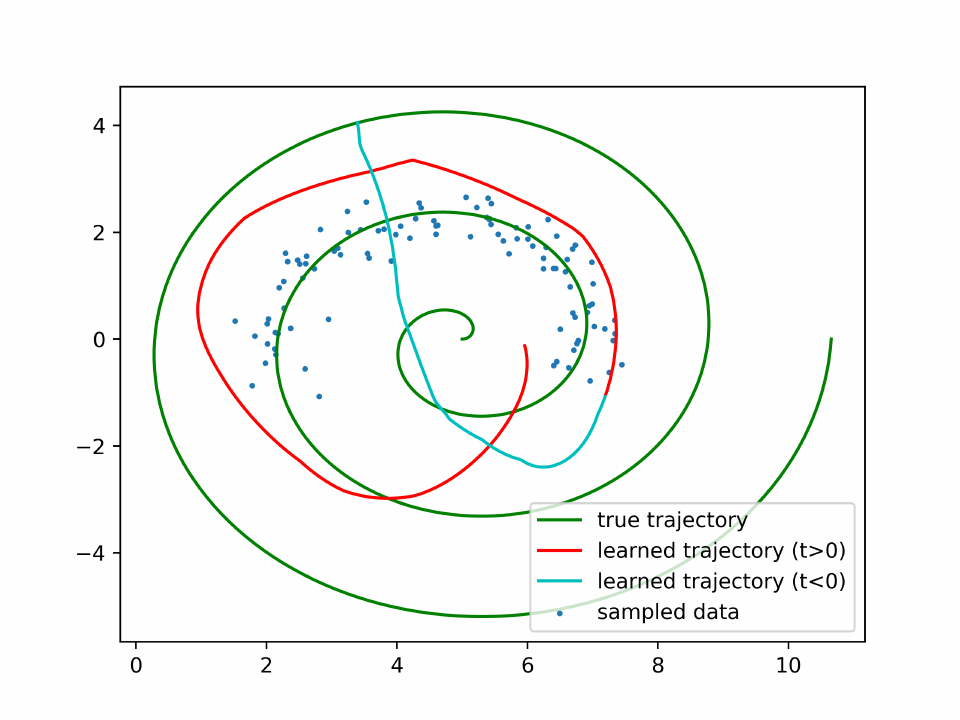}
    \caption{Latent ODE-LSTM+GC}
    \label{fig:spiral100c}
    \end{subfigure}
    \caption{Visual performance of Latent ODE-RNN (a), Latent ODE-LSTM (b) and Latent ODE-LSTM+GC-Gradient Clipping (c) in the task of reconstruction (best fit to the true trajectory for the sampled data time points) and extrapolation (backwards in time, $t< 0$, shown in blue, and forward in time, $t > 0$, in red) of counterclockwise (top) and clockwise (bottom) spirals, after training with datasets with $N=100$. }
    \label{fig:spiral100}
\end{figure}

\begin{figure}[H]
    \begin{subfigure}[b]{.33\textwidth}
    \includegraphics[width=\textwidth]{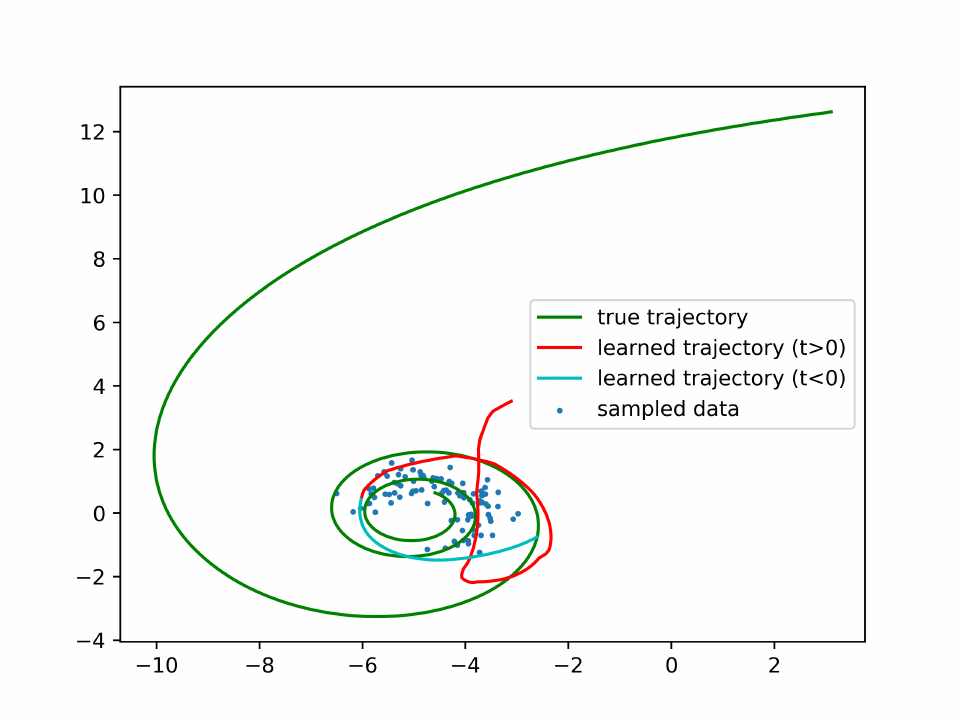} \\ \includegraphics[width=\textwidth]{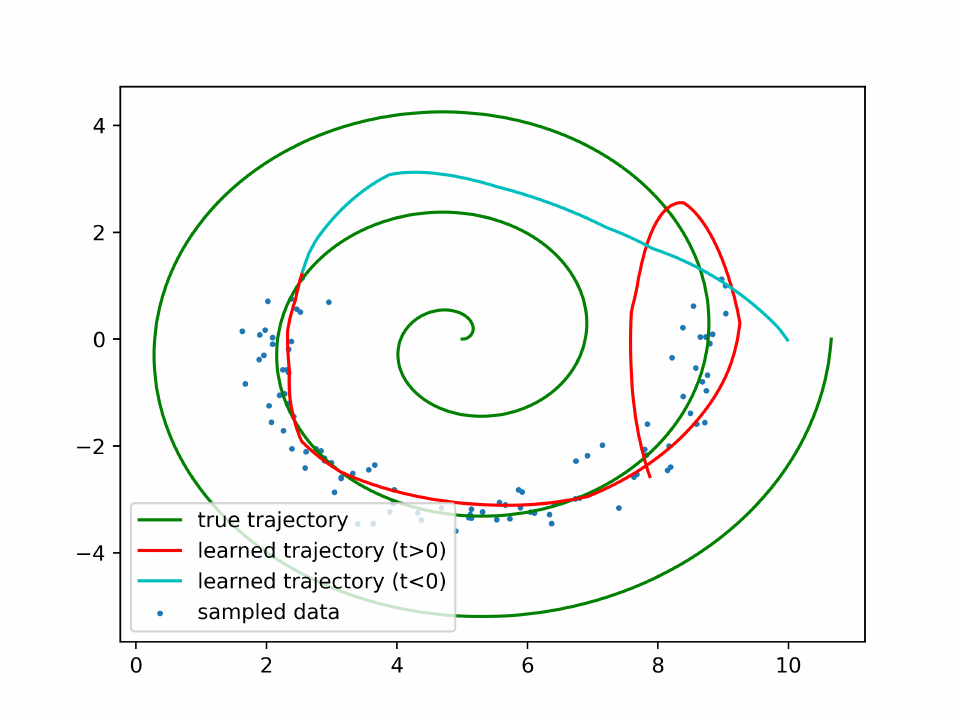}
    \caption{Latent ODE-RNN}
    \label{fig:spiral50a}
    \end{subfigure}
    \begin{subfigure}[b]{.33\textwidth}
    \includegraphics[width=\textwidth]{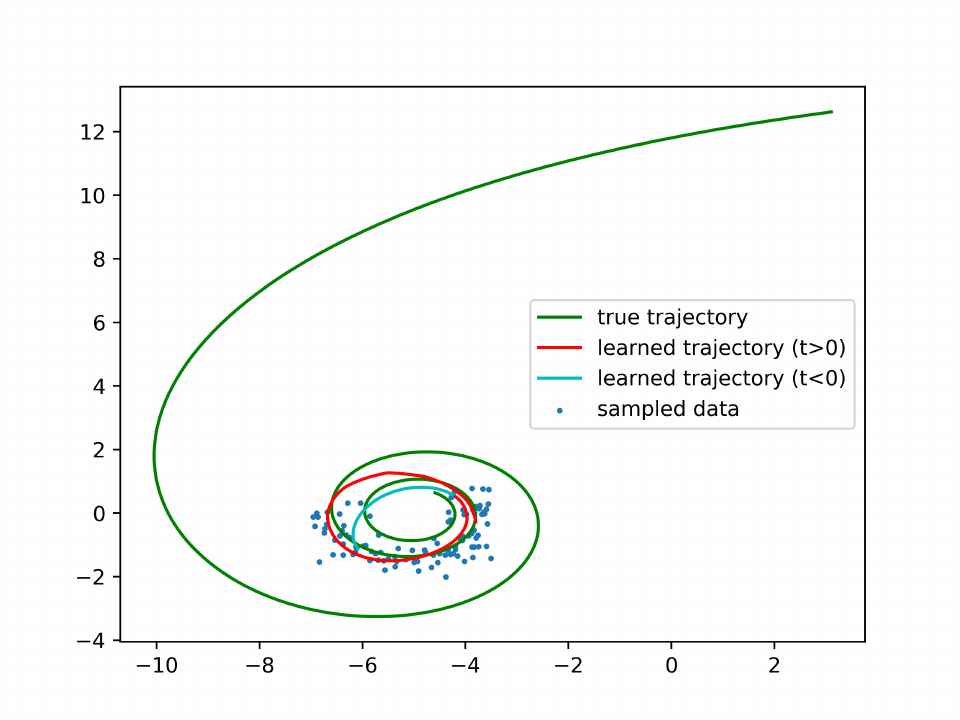} \\ \includegraphics[width=\textwidth]{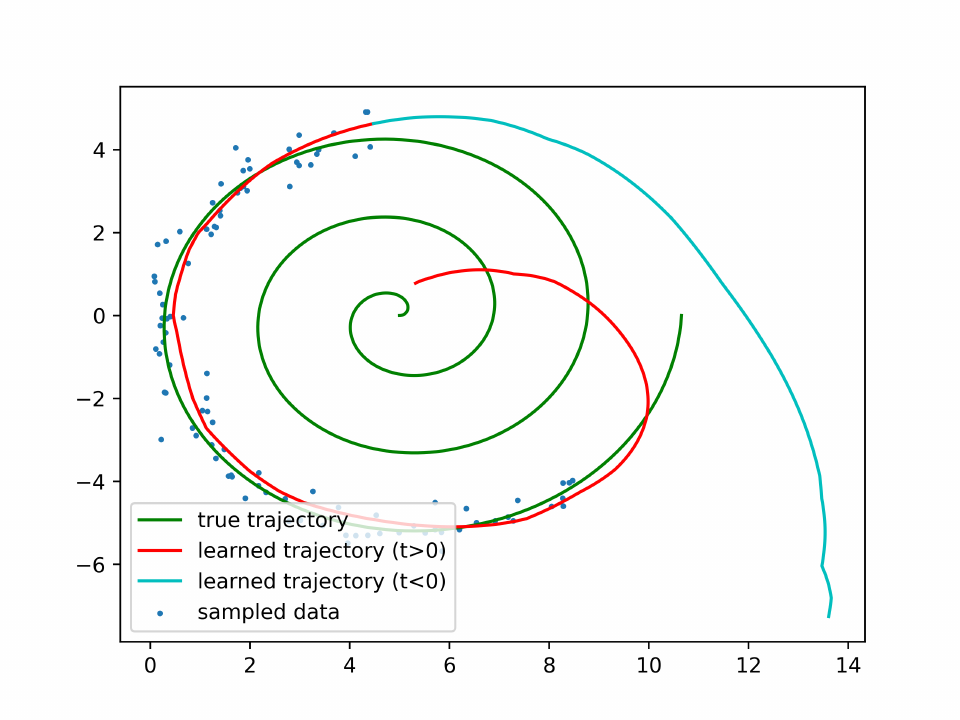}
    \caption{Latent ODE-LSTM}
    \label{fig:spiral50b}
    \end{subfigure}
    \begin{subfigure}[b]{.33\textwidth}
    \includegraphics[width=\textwidth]{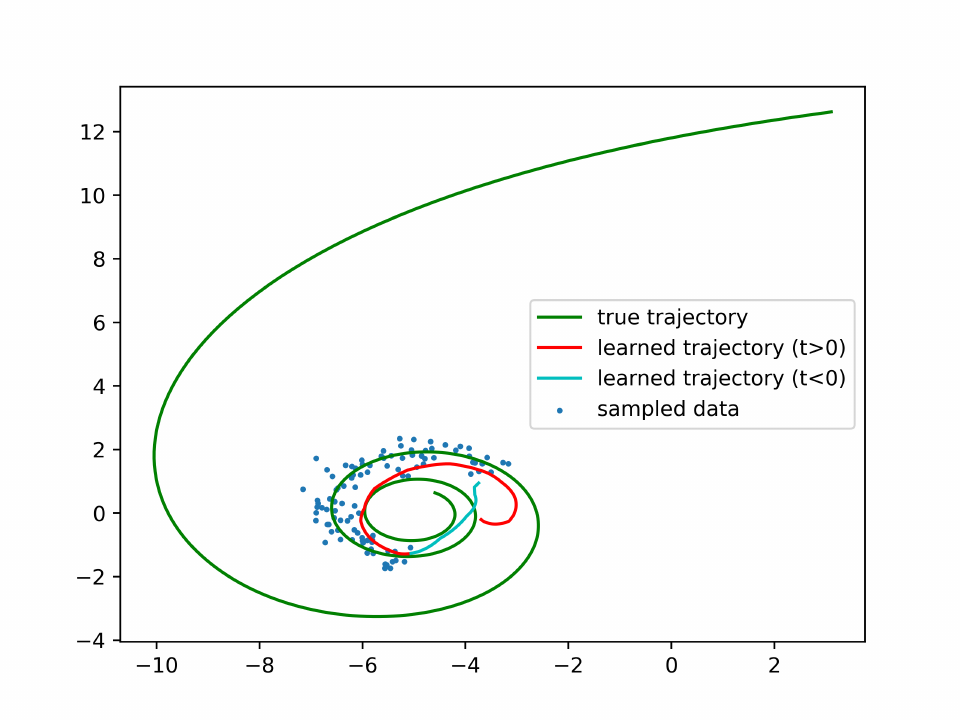} \\ \includegraphics[width=\textwidth]{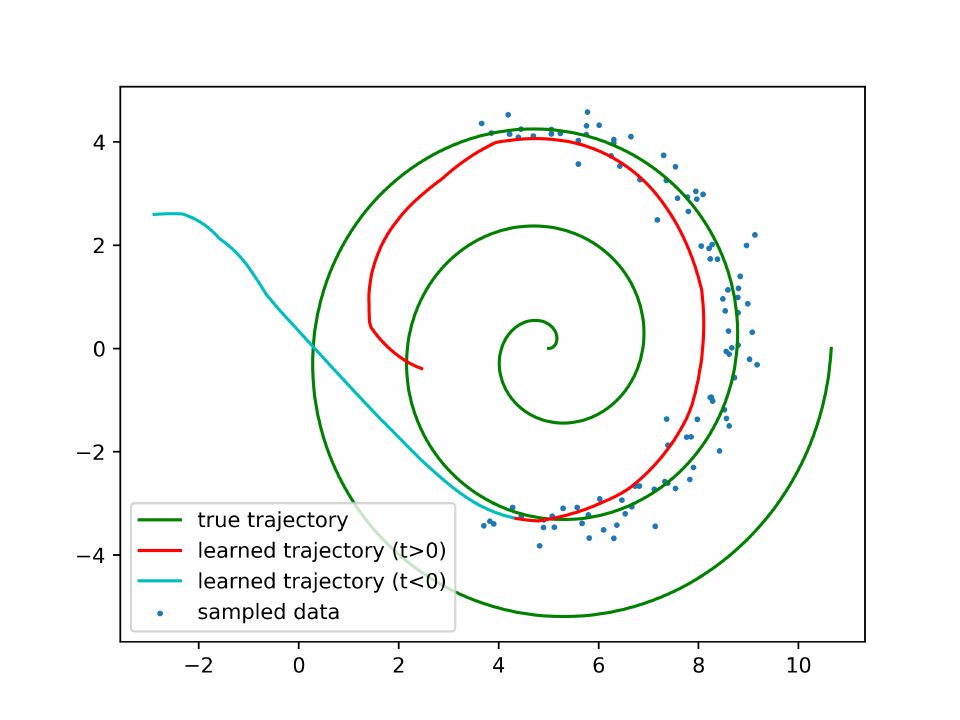}
    \caption{Latent ODE-LSTM+GC}
    \label{fig:spiral50c}
    \end{subfigure}
    \caption{Visual performance of Latent ODE-RNN (a), Latent ODE-LSTM (b) and Latent ODE-LSTM+GC-Gradient Clipping (c) at the task of reconstruction (best fit to the true trajectory for the sampled data time points) and extrapolation (backwards in time, $t< 0$, shown in blue, and forward in time, $t > 0$, in red) of counterclockwise (top) and clockwise (bottom) spirals, after training with datasets with $N=50$. }
    \label{fig:spiral50}
\end{figure}

\begin{figure}[H]
    \begin{subfigure}[b]{.33\textwidth}
    \includegraphics[width=\textwidth]{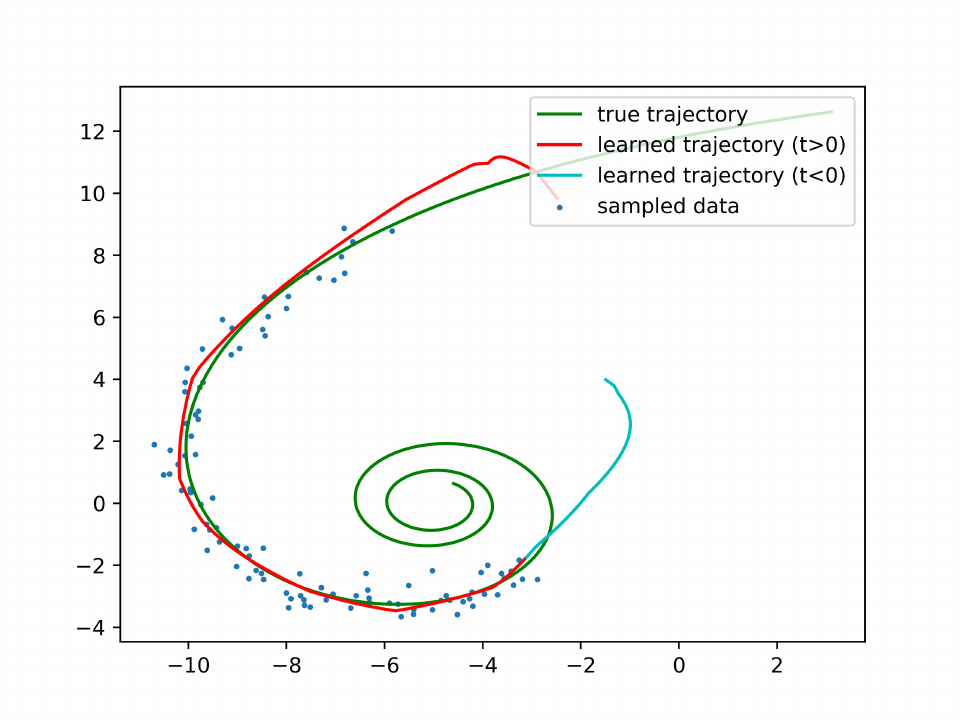} \\ \includegraphics[width=\textwidth]{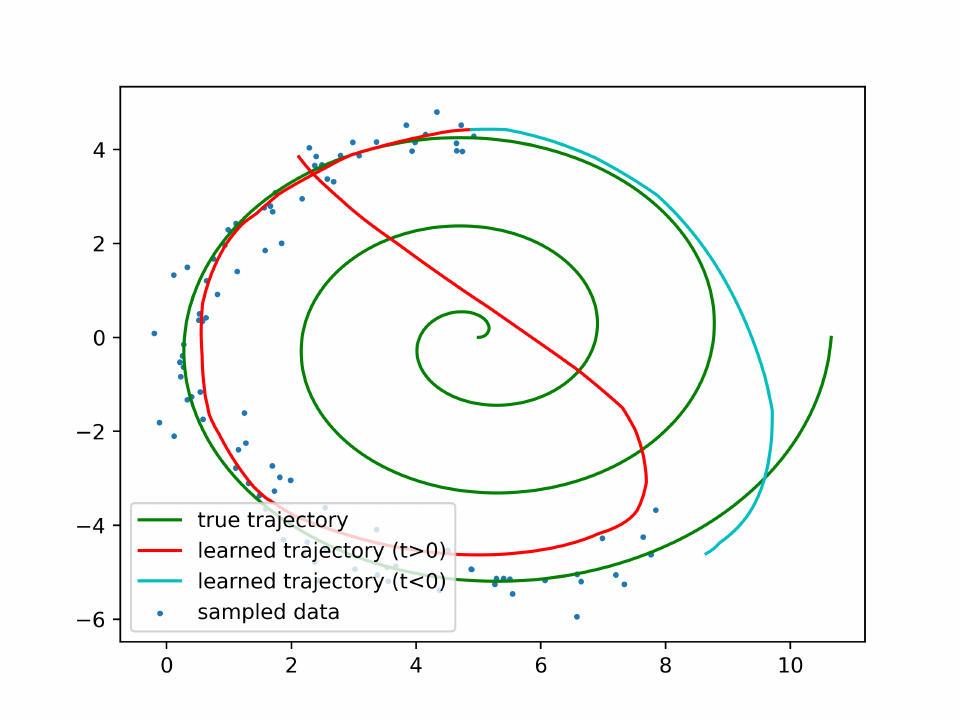}
    \caption{Latent ODE-RNN}
    \label{fig:spiral30a}
    \end{subfigure}
    \begin{subfigure}[b]{.33\textwidth}
    \includegraphics[width=\textwidth]{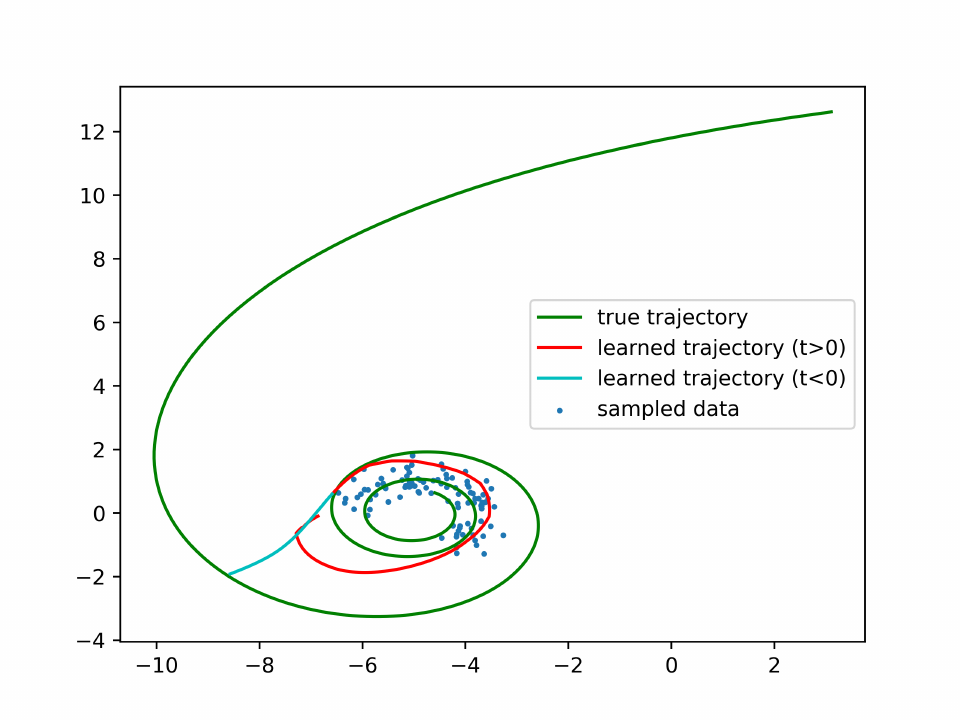} \\ \includegraphics[width=\textwidth]{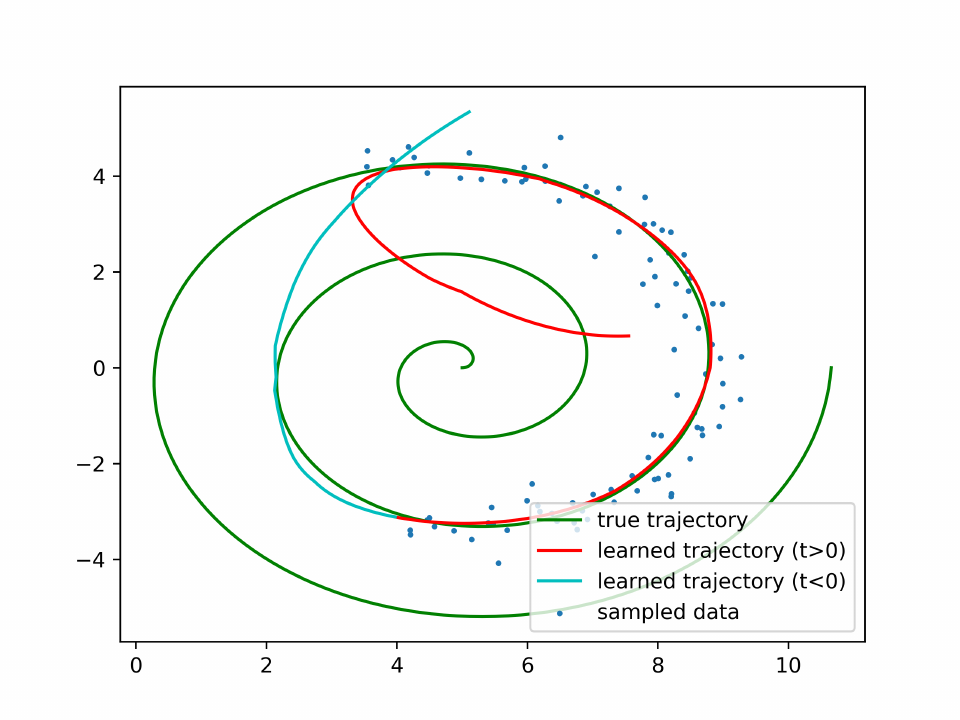} 
    \caption{Latent ODE-LSTM}
    \label{fig:spiral30b}
    \end{subfigure}
    \begin{subfigure}[b]{.33\textwidth}
    \includegraphics[width=\textwidth]{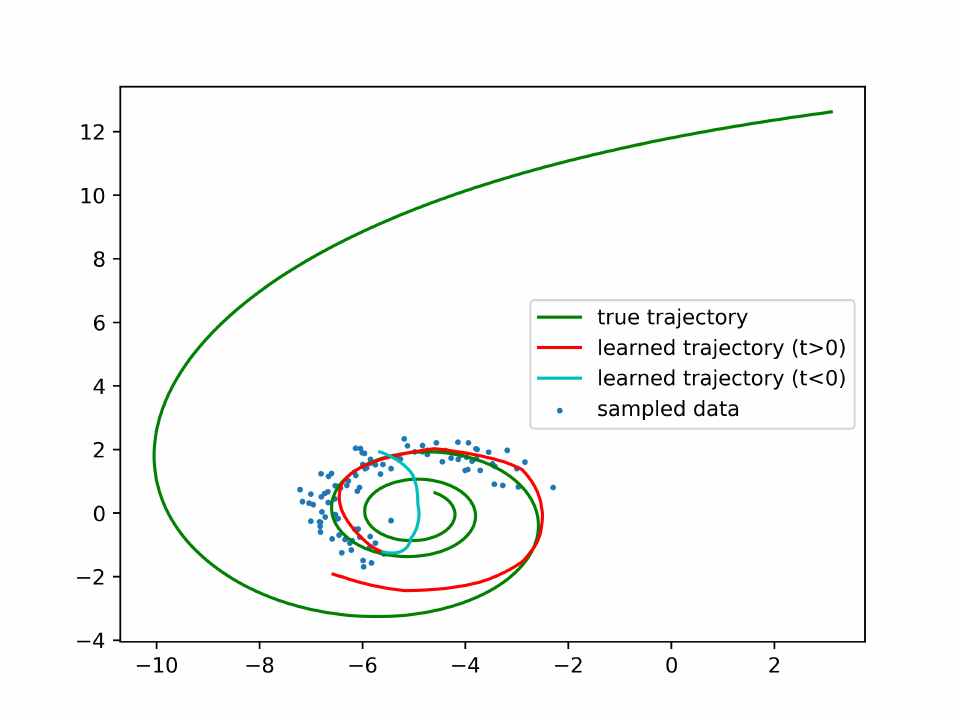} \\ \includegraphics[width=\textwidth]{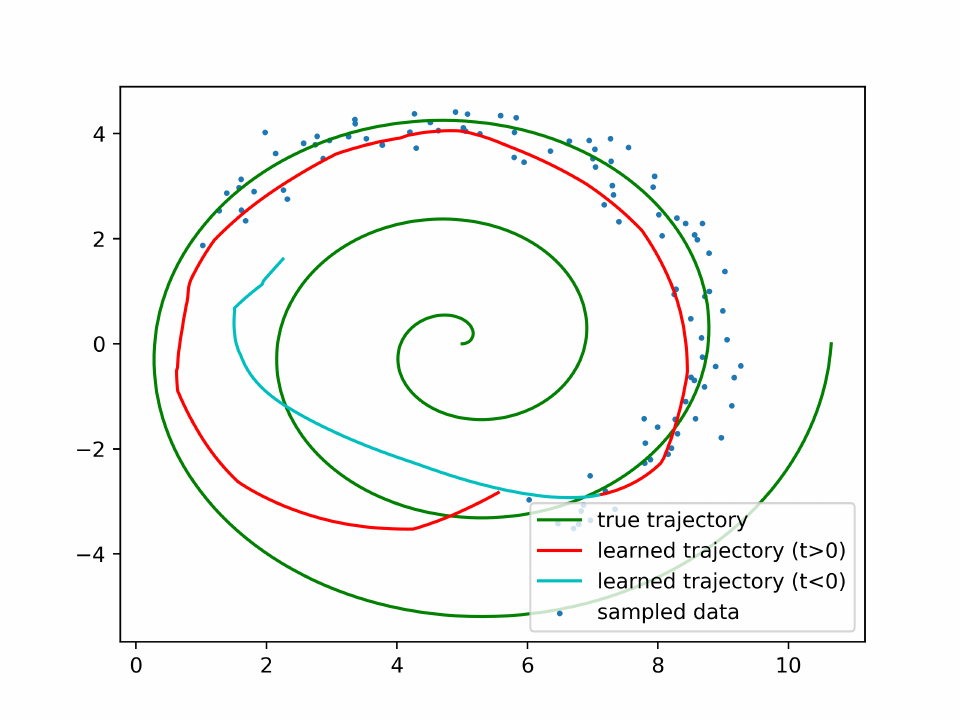}
    \caption{Latent ODE-LSTM+GC}
    \label{fig:spiral30c}
    \end{subfigure}
    \caption{Visual performance of Latent ODE-RNN (a), Latent ODE-LSTM (b) and Latent ODE-LSTM+GC-Gradient Clipping (c) at the task of reconstruction (best fit to the true trajectory for the sampled data time points) and extrapolation (backwards in time, $t< 0$, shown in blue, and forward in time, $t > 0$, in red) of counterclockwise (top) and clockwise (bottom) spirals, after training with datasets with $N=30$. }
    \label{fig:spiral30}
\end{figure}

Testing was performed using $D_{N=500}$ for seen (reconstruction) and unseen time points (extrapolation) by each model.
The results obtained in the task of reconstructing and extrapolating bidirectional spirals for Latent ODE-RNN, Latent ODE-LSTM and Latent ODE-LSTM+GC are shown in Figure \ref{fig:spiral250} for $N=250$, Figure \ref{fig:spiral100} for $N=100$, Figure \ref{fig:spiral50} for $N=50$ and Figure \ref{fig:spiral30} for $N=30$.

For models trained with $D_{N=250}$ (Figure \ref{fig:spiral250}), one can see from the results that the counterclockwise spiral (top of Figure \ref{fig:spiral250}) reconstructions are far from the sampled data points. The extrapolation backward in time, $t< 0$, is completely off the mark and shows no resemblance to spiral dynamics. The Latent ODE-LSTM model stands out from the others when extrapolated forward in time, $t > 0$, and shows dynamics that are close to the target, albeit a little shift from the true trajectory. The results for the clockwise spiral (bottom of Figure \ref{fig:spiral250}) are better. Latent ODE-LSTM is the best model for reconstruction, but cannot extrapolate for $t< 0$. Latent ODE-LSTM+GC had the best performance on the extrapolation task.

For models trained with $D_{N=100}$, Figure \ref{fig:spiral100}, Latent ODE-LSTM+GC showed better extrapolation of the counterclockwise spiral (top of Figure \ref{fig:spiral100}). In the reconstruction task, Latent ODE-RNN and Latent ODE-LSTM+GC show similar performance.
Looking at the clockwise spiral results (bottom of Figure \ref{fig:spiral100}), we can see that all models perform worse than the models trained with $D_{N=250}$.

For the models trained with $D_{N=50}$, the three models had difficulty reconstructing and extrapolating the counterclockwise spirals, Figure \ref{fig:spiral50} top.
As can be seen for the models trained with $D_{N=250}, D_{N=100}$ in Figure \ref{fig:spiral250} and Figure \ref{fig:spiral100}, Latent ODE-LSTM has the best performance in reconstructing the clockwise spirals and Latent ODE-LSTM+GC is the best at extrapolation. Latent ODE-RNN's extrapolation is worst at both tasks, Figure \ref{fig:spiral50} bottom.

For the models trained with $D_{N=30}$, from Figure \ref{fig:spiral30} we can see that the performance of the three models in reconstruction (Figure \ref{fig:spiral30} bottom) remains similar to experiments with models trained with more time points, $D_{N=50},D_{N=100},D_{N=250}$, with Latent ODE-LSTM performing best on this task.
When extrapolating, forward and backward in time, Latent ODE-LSTM+GC shows a better fit to the dynamics of the spiral for longer time periods (end of sampled data points in Figure \ref{fig:spiral30}). We note that the Latent ODE-RNN and Latent ODE-LSTM models interrupt the dynamics of the spiral immediately after the end of the sampled points for $t > 0$.

\subsection{Quantitative evaluation}

\subsubsection{Experimental conditions}
The VAE models for the qualitative evaluation consist of: ODE-LSTM (or ODE-RNN) encoder with a hidden layer of $4$ neurons, a Neural ODE with a hidden layer of $25$ neurons and an output layer of $4$ neurons; and a decoder with a Neural ODE with $1$ hidden layer with $25$ neurons and an output network with $1$ hidden layer and $256$ neurons.
The Neural ODEs are solved with Runge-Kutta method of order 5 of Dormand-Prince-Shampine with an adaptive step size.
For training, we chose the Adam optimiser with $0.0005$ learning rate. 

To evaluate the performance of the models in extrapolating short- and long-term sequences, 4 pairs of input and prediction sequences of different lengths were used: 7 days, 15 days, 30 days and an input of 365 days to predict the next 60 days.

Each combination of model and sequence length was run $3$ times ($N=3$), to account for the randomness in the models during training. Each model trained for $50$ (daily climate series data) and $100$ (DJIA 30 stock time series) epochs.
The mean square error and standard deviation of the test data were determined.

\subsubsection{Daily Climate Time Series Data}

In this numerical experiment, we have investigated the ability of the models to extrapolate the weather forecast for Delhi, India using 4 parameters: Date, Mean Temperature, Humidity, Wind Speed, and Mean Air Pressure.
The training dataset used provides a training time series with one value per day between January 1, 2013 and January 1, 2017, resulting in 1462 values. The test data consists of 114 daily data points between January 1 2017 and the 24th of April 2017 \cite{climate}.

From this dataset we construct $4$ training datasets each with a different sequence length $N={7,15,30,365}$ selected consecutively, denoted by $D_{N=7},D_{N=15},D_{N=30},D_{N=365}$.

Three models were trained, namely Latent ODE-RNN, Latent ODE-LSTM and Latent ODE-LSTM+GC, and their performance was evaluated by computing the mean of MSE and its standard deviation for the test set, Table \ref{table:climate}.

Latent ODE-LSTM showed a lower MSE in all experiments with different sequence lengths (seen/predict).
In general, the models show similar performance.

\begin{table}[H]
\centering
\resizebox{\textwidth}{!}{%
\begin{tabular}{llll}
\cline{2-4}
             &                      &   Mean Squared Error                  &                                   \\ \hline
seen/predict & Latent ODE-RNN       & Latent ODE-LSTM     & Latent ODE-LSTM+GC \\ \hline
$D_{N=7}$/7          & 0.228$\pm$     0.001 & 0.227 $\pm$  0.001  & 0.230 $\pm$ 0.001                 \\
$D_{N=15}$/15        & 0.235$\pm$  0.002    & 0.232 $\pm$ 0.002   &     0.235 $\pm$ 0.001 \\
$D_{N=30}$/30        & 0.216$\pm$  0.001    & 0.212$\pm$    0.002 &  0.233$\pm$ 0.001 \\
$D_{N=365}$/60       & 0.220$\pm$    0.001  & 0.219$\pm$  0.002   & 0.226 $\pm$ 0.003                 \\ \hline
\end{tabular}%
}
\caption{Mean Squared Error, for Latent ODE-RNN and Latent ODE-LSTM with and without norm gradient, for the test set of daily climate time series for input and forecast sequences of different lengths. (mean $\pm$ standard deviation, $N=3$).}
\label{table:climate}
\end{table}

\subsubsection{DJIA 30 Stock Time Series}

In this experiment, we evaluated the performance of the models in extrapolating stock market data. The dataset used is irregularly sampled and yields 1 parameter vector per day, between January 3, 2006 and December 29, 2017, with 6 parameters: date, price of the stock when the stock market opened, highest and lowest price reached on that day, number of shares traded (discarded in our experiments) and the stock's ticker name \cite{djia}. Although data from 29 DJIA companies are available, only one was used. In total, 3019 data points are available to train and test the models, and a split of 75/25 was used.

From this dataset we construct $4$ training datasets each with a different sequence length $N={7,15,30,365}$ selected consecutively, denoted by $D_{N=7},D_{N=15},D_{N=30},D_{N=365}$.

Three models were trained, namely Latent ODE-RNN, Latent ODE-LSTM and Latent ODE-LSTM+GC, and their performance was evaluated by computing the mean MSE and its standard deviation for the test set, Table \ref{table:DJIA}.

From Table \ref{table:DJIA}, we can see that Latent ODE-RNN is slightly better in all experiments, except for the model trained with $D_{N=365}$. As expected, the Latent ODE-LSTM models show better results for training with $D_{N=365}$, since the LSTM architecture is devoted to long-term sequences.

The models clearly show the difficulty in gradually predicting longer sequences, ranging from an MSE on the order of $10^{-2}$ for a short-term prediction of $7$ days to $10^{-1}$ for a longer-term prediction of $60$ days.

\begin{table}[H]
\resizebox{\textwidth}{!}{
\begin{tabular}{llll}
\cline{2-4}
             &                      &  Mean Squared Error                    &                          \\ \hline
seen/predict & Latent ODE-RNN       & Latent ODE-LSTM      & Latent ODE-LSTM+GC \\ \hline
$D_{N=7}$/7          & 0.025  $\pm$   0.002 & 0.044 $\pm$    0.005 & 0.077$\pm$    0.006      \\
$D_{N=15}$/15        & 0.030 $\pm$    0.013 & 0.069 $\pm$ 0.019    & 0.125$\pm$ 0.039         \\
$D_{N=30}$/30        & 0.096$\pm$   0.040   & 0.161 $\pm$  0.021   & 0.198 $\pm$ 0.006        \\
$D_{N=365}$/60       & 0.245$\pm$ 0.012     & 0.232$\pm$   0.046   & 0.233 $\pm$ 0.002        \\ \hline
\end{tabular}%
}
\caption{Mean Squared Error of Latent ODE-RNN and Latent ODE-LSTM with and without Norm Gradient Clipping, on DJIA 30 Stock Time Series test set for sequences with different input and prediction lengths. (mean $\pm$ standard deviation, $N=3$).}
\label{table:DJIA}
\end{table}

\section{Conclusions} \label{sec:conclusion}

In this paper, to overcome the vanishing and exploding gradients during training of Latent ODE-RNN, we proposed Latent ODE-LSTM. This architecture is a Variational Autoencoder with an ODE-LSTM encoder, in which the state transitions between LSTM cells are given by a Neural ODE, and a Neural ODE decoder.
Using a LSTM in the encoder our architecture is more flexible and further retains the information given by past values.
We proved that Latent ODE-LSTM network mitigates the vanishing gradient problem. 
Latent ODE-LSTM does not solve the exploding gradient problem, and a proof has been derived. 

To limit the growth of the gradients, the norm gradient clipping strategy was embedded in the model.
This strategy provides explicit control over the gradients by rescaling them when their norm is greater than a predefined threshold.

To evaluate the performance of the Latent ODE-LSTM models, we performed two types of evaluation, qualitative and quantitative, using a Latent ODE-RNN baseline.
The qualitative approach was to visually test the reconstruction and extrapolation of the dynamics of bidirectional clockwise and counterclockwise spirals using synthetic datasets with irregularly sampled time points and gradually sparser data.
The results show that Latent ODE-LSTM has the best performance in reconstruction and that sparser data does not affect performance. Latent ODE-LSTM + GC has the best performance in extrapolation backward, $t< 0$ and forward, $t > 0$, in time.

A qualitative approach was performed by considering two real-life time series, one sampled regularly and one sampled irregularly. The data of these datasets were split into smaller sequences to test both short- and longer-term predictions.
The numerical experiments showed that the models have similar performances. Contrary to the Latent ODE-RNN model, the Latent ODE-LSTM model mitigates the vanishing and exploding gradients problem. Furthermore, Latent ODE-LSTM + GC does not suffer from the exploding gradient problem.

There are several directions that can be taken in the future. 
The choice of the numerical solver of the Neural ODE encoder and decoder has proved to be challenging. In the future, it would be important to study the correlation between the hyperparameters of the architecture (number of layers, neurons, activation functions) and the characteristics of the dataset, as well as the performance of the ODE solver and the best choice of the numerical method.

\section{Acknowledgements}
The authors acknowledge the funding by Fundação para a Ciência e Tecnologia (Portuguese Foundation for Science
and Technology) through CMAT projects UIDB/00013/2020 and UIDP/00013/2020.
C. Coelho would like to thank FCT for the funding through the scholarship with reference 2021.05201.BD.


\bibliographystyle{ieeetr}

\newpage
\appendix

\section{Vanishing and exploding gradients problem in RNN-based networks} \label{sec:gradient}

In this section we show that Latent ODE-RNNs suffer from the vanishing and exploding gradients problem. To prove this, we first show that RNNs suffer from the vanishing and exploding problem \cite{bengioLearningLongtermDependencies1994,pascanuDifficultyTrainingRecurrent2013}, and then that ODE-RNNs, Latent ODEs and Latent ODE-RNNs inherit this problem.

Let $l(\hat{y}_i)$ be the loss function value at time step $i$, then backpropagation through time (BPTT) is used to optimise the parameters $\theta$ of an RNN, with the total loss given by the sum of losses of all time steps,

$$\mathcal{L}_\theta = \dfrac{1}{N} \sum_{i=1}^N l(\hat{y}_i).$$

To find the parameters $\theta$ that minimise the loss $\mathcal{L_\theta}$, the gradients are then calculated , Equation \ref{eq:rnn1},

\begin{equation}
\label{eq:rnn1}
\dfrac{\partial \mathcal{L}}{\partial \theta} = \dfrac{1}{N} \sum_{i=1}^N \dfrac{\partial l(\hat{y}_i)}{\partial \theta}.
\end{equation}
{\color{black}The term $\dfrac{\partial l(\hat{y}_i)}{\partial \theta}$ is a sum of products that gives the gradients at time step $i$ (considering all the contributions of the previous time steps $k$, where $k<i$) \eqref{eq:rnn2},}

\begin{equation}
\label{eq:rnn2}
\dfrac{\partial l(\hat{y}_i)}{\partial \theta}= \sum_{k=1}^i \left( \dfrac{\partial l(\hat{y}_i)}{\partial \hat{y}_i} \dfrac{\partial \hat{y}_i}{\partial h_i} \dfrac{\partial h_i}{\partial h_{k}} \dfrac{\partial^+ h_k}{\partial \theta} \right),
\end{equation}

where $\dfrac{\partial^+ h_k}{\partial \theta}$ is the \emph{immediate} partial derivative of $h_k$ with respect to the parameters $\theta$ \cite{pascanuDifficultyTrainingRecurrent2013}.

\subsection{RNNs and the vanishing/exploding gradient problem} \label{proof:RNN}

As seen in Section \ref{sec:background}, RNNs are neural networks with feedback loops that, when unfolded, can be viewed as a Feed-forward Neural Network with $N$ layers, where each layer is an RNN cell.

Since the layers are copies of the same RNN cell, the parameters $\theta=(\textbf{w}_{\text{input}},\textbf{w}_{\text{feedback}},b, \textbf{w}_{\text{output}}, \textbf{b}_{\text{output}})$ are shared across the network depth, so that the computation of the hidden states $h_{i}, \, i\in(1 \dots N)$ consists of multiple instances of the same values \cite{pascanuDifficultyTrainingRecurrent2013}.

The term $\dfrac{\partial h_i}{\partial h_k}$ is a chain of products of all the hidden states that contribute to the hidden state at time step $i$ (see also the RNN update \eqref{eq:bck_rnn1}),

\begin{equation}
\label{eq:rnn3}
\dfrac{\partial h_i}{\partial h_k} = \left( \prod_{i\geq j>k} \dfrac{\partial h_{j}}{\partial h_{j-1}} \right) = \left( \prod_{i\geq j>k} \textbf{w}_{\text{feedback}}  diag(\sigma'(  \textbf{w}_{\text{feedback}} h_{j-1}+  \textbf{w}_{\text{input}} x_j + b)) \right).
\end{equation}
The chain of products in\eqref{eq:rnn3} can be rewritten as a power of $i-k$ terms, where $diag()$ is the transformation of the vector of derivatives of the activation function $\sigma$ into a diagonal matrix \cite{pascanuDifficultyTrainingRecurrent2013},

\begin{equation}
\label{eq:rnn6}
\left( \textbf{w}_{\text{feedback}}  diag(\sigma'(\textbf{w}_{\text{feedback}} h_{j-1}  +  \textbf{w}_{\text{input}} x_j + b ))\right)^{i-k}
\end{equation}

with

\begin{equation}
\begin{array}{rcl}
\label{eq:rnn6.0}
 \left \lVert \dfrac{\partial h_j}{\partial h_{j-1}} \right \rVert=\left \lVert \textbf{w}_{\text{feedback}}  diag(\sigma'(  \textbf{w}_{\text{feedback}} h_{j-1}+  \textbf{w}_{\text{input}} x_j + b )) \right \rVert \leq \\ \left \lVert \textbf{w}_{\text{feedback}} \right \rVert  \left \lVert diag(\sigma'( \textbf{w}_{\text{feedback}} h_{j-1} +   \textbf{w}_{\text{input}} x_j+ b )) \right \rVert.
\end{array}
\end{equation}

Let $\sigma$ be any activation function such that $\sigma'$ is its derivative, bounded by the absolute value $\gamma$ and therefore $\lVert diag(\sigma'(.)\rVert \leq \gamma$. Thus $\left \lVert diag(\sigma'( \textbf{w}_{\text{feedback}} h_{j-1} + \textbf{w}_{\text{input}} x_j + b)) \right \rVert \leq \gamma$ \cite{pascanuDifficultyTrainingRecurrent2013}.

\subsubsection{Vanishing gradient}

The vanishing gradient problem occurs when the norm of gradients decreases exponentially and tends to zero during the training of a NN. This prevents the optimisation process from learning the best parameters $\theta$ that minimise the loss function \cite{bengioLearningLongtermDependencies1994, pascanuDifficultyTrainingRecurrent2013}.

Following the proof in \cite{pascanuDifficultyTrainingRecurrent2013}, we take $\lambda$ to be the absolute value of the largest eigenvalue of the feedback matrix, $\textbf{w}_{\text{feedback}}$, and for all $j$, it is sufficient to have

\begin{equation*}
\forall j, \left \lVert  \dfrac{\partial h_j}{\partial h_{j-1}}  \right \rVert < \dfrac{1}{\gamma} \gamma < 1.
\end{equation*}

For all $j$, taking $\eta \in \mathbb{R}$ so that $\eta < 1$ comes $\left \lVert  \dfrac{\partial h_j}{\partial h_{j-1}}  \right \rVert \leq \eta < 1$. By induction over $j$ we obtain,

\begin{equation}
\label{eq:eta1}
    \dfrac{\partial l(\hat{y}_i)}{\partial \hat{y}_i} \dfrac{\partial \hat{y}_i}{\partial h_i} 
 \left( \prod_{i \geq j \geq k} \dfrac{\partial h_j}{\partial h_{j-1}} \right) \leq  \dfrac{\partial l(\hat{y}_i)}{\partial \hat{y}_i} \dfrac{\partial \hat{y}_i}{\partial h_i}\eta^{i-k}.
\end{equation}

As $\eta < 1$ it follows from \ref{eq:eta1} that in the presence of long-term sequences of data (for which $i-k$ is large) the term $\eta^{i-k}$ decreases exponentially fast to zero leading to the vanishing of the gradients \cite{pascanuDifficultyTrainingRecurrent2013}.

\subsubsection{Exploding gradient}

The problem of exploding gradients arises when the norm of gradients increases exponentially towards infinity \cite{bengioLearningLongtermDependencies1994, pascanuDifficultyTrainingRecurrent2013}. This makes the optimisation steps very large and it is impossible to reach an optimal point.

Inverting the vanishing gradient proof, it is necessary to have

\begin{equation*}
\forall j, \left \lVert  \dfrac{\partial h_j}{\partial h_{j-1}}  \right \rVert > \dfrac{1}{\gamma} \gamma > 1
\end{equation*}

Taking $\eta \in \mathbb{R}$, for all $j$, so that $\eta>1$ comes $\left \lVert  \dfrac{\partial h_j}{\partial h_{j-1}}  \right \rVert \geq \eta > 1$. By induction over $j$ we obtain,

\begin{equation}
\label{eq:eta2}
    \dfrac{\partial l(\hat{y}_i)}{\partial \hat{y}_i} \dfrac{\partial \hat{y}_i}{\partial h_i} 
 \left( \prod_{i \geq j \geq k} \dfrac{\partial h_j}{\partial h_{j-1}} \right) \geq  \dfrac{\partial l(\hat{y}_i)}{\partial \hat{y}_i} \dfrac{\partial \hat{y}_i}{\partial h_i}\eta^{i-k}.
\end{equation}

As $\eta > 1$ it follows from \ref{eq:eta2} that in the presence of long-term sequences of data (for which $i-k$ is large) the term $\eta^{i-k}$ increases exponentially fast leading to the explosion of the gradients \cite{pascanuDifficultyTrainingRecurrent2013}.

\subsection{ODE-RNNs and the vanishing/exploding gradient problem} \label{proof:ODE-RNN}

As seen in Section \ref{sec:background}, the difference between RNN and ODE-RNN is that to calculate the hidden state $h_{i}$, an intermediate state $h'_{i}$ is computed by solving an ODE (when solving the NN adjusted ODE $f_\theta$) using the previous state, $h_{i-1}$, (given by an RNN) as initial condition (see Figure \ref{fig:ODE-RNN}),

\begin{equation*}
h_{i} = \sigma(\textbf{w}_{\text{feedback}} \underbrace{ODESolve(f_\theta,h_{i-1},(t_{i-1},t_i))}_{h'_{i}}   + \textbf{w}_{\text{input}}  x_{i}+b).
\end{equation*}
To find the parameters $\theta$ in the ODE-RNN that minimise the loss $\mathcal{L}_{\theta}$, the term $\dfrac{\partial h_i}{\partial h_k}$ in \eqref{eq:rnn2} must take into account the intermediate state $h'_{i}$. Converting $\dfrac{\partial h'_j}{\partial h_{j-1}}$ into a diagonal matrix, $\dfrac{\partial h_i}{\partial h_k}$ is given by:

\begin{equation}
\begin{array}{rcl}
\label{eq:odernn5}
\dfrac{\partial h_i}{\partial h_k} &=& \left( \prod\limits_{i\geq j>k} \dfrac{\partial h_{j}}{\partial h'_{j}} \,diag\left(\dfrac{\partial h'_{j}}{\partial h_{j-1}}\right) \right) 
\\  &=& \left( \prod\limits_{i\geq j>k} \textbf{w}_{\text{feedback}}  diag(\sigma'( \textbf{w}_{\text{feedback}} h_{j-1} +  \textbf{w}_{\text{input}} x_j + b))   diag \left( \dfrac{\partial h'_j}{\partial h_{j-1}} \right) \right).
\end{array}
\end{equation}

As in RNNs, the term $\dfrac{\partial h_i}{\partial h_k}$ in ODE-RNNs is a chain of products, that now also has the partial derivative of the ODE solver, $\dfrac{\partial h'_{j}}{\partial h_{j-1}}$. Thus, \eqref{eq:odernn5} can be rewritten as a power of $i-k$ terms,

\begin{equation}
\label{eq:odernn6}
\left( \textbf{w}_{\text{feedback}}  diag(\sigma'(\textbf{w}_{\text{feedback}} h_{j-1}  +  \textbf{w}_{\text{input}} x_j + b))  diag \left( \dfrac{\partial h'_j}{\partial h_{j-1}} \right) \right)^{i-k}
\end{equation}

with

\begin{equation}
\begin{array}{rcl}
\label{eq:odernn6.1}
 \left \lVert \dfrac{\partial h_j}{\partial h'_j} diag\left(\dfrac{\partial h'_j}{\partial h_{j-1}}\right) \right \rVert=\left \lVert \textbf{w}_{\text{feedback}}  diag(\sigma'(\textbf{w}_{\text{feedback}} h_{j-1}  +  \textbf{w}_{\text{input}} x_j + b))  diag \left(\dfrac{\partial h'_j}{\partial h_{i-j}} \right) \right \rVert \leq \\ \left \lVert \textbf{w}_{\text{feedback}}  diag(\sigma'(\textbf{w}_{\text{feedback}} h_{j-1}  + \textbf{w}_{\text{input}} x_j  + b)) \right \rVert  \left \lVert diag \left( \dfrac{\partial h'_j}{\partial h_{j-1}} \right) \right \rVert
\end{array}
\end{equation}

where the first term, $\left \lVert \textbf{w}_{\text{feedback}} diag(\sigma'( \textbf{w}_{\text{feedback}} h_{j-1} + \textbf{w}_{\text{input}} x_j + b)) \right \rVert$ is the same as in \eqref{eq:rnn6}, while the second term $\left \lVert diag \left( \dfrac{\partial h'_j}{\partial h_{j-1}} \right) \right \rVert$ appears due to the ODE solver used to compute the hidden states $h'_j$ in the RNN update \eqref{eq:bckLatentODERNN2}.

\subsubsection{Vanishing gradient}

Consider $\lambda$ and $\lambda'$ as the absolute values of the largest eigenvalue of the feedback matrix, $\textbf{w}_{\text{feedback}}$, and the Jacobian matrix $ diag\left(\dfrac{\partial h'_j}{\partial h_{j-1}}\right)$, respectively.
Following the vanishing gradient proof done for RNNs, \ref{proof:RNN}, it is sufficient to have $\lambda \lambda' < \dfrac{1}{\gamma}$ for

\begin{equation*}
\forall j, \left \lVert  \dfrac{\partial h_j}{\partial h'_j} diag\left(\dfrac{\partial h'_j}{\partial h_{j-1}}\right)  \right \rVert < \dfrac{1}{\gamma} \gamma < 1.
\end{equation*}

For all $j$, taking $\eta \in \mathbb{R}$ such that $\eta<1$ comes $\left \lVert  \dfrac{\partial h_j}{\partial h'_j}  diag\left(\dfrac{\partial h'_j}{\partial h_{j-1}}\right)  \right \rVert \leq \eta < 1$. By induction over $j$ we obtain

\begin{equation}
\label{eq:eta3}
    \dfrac{\partial l(\hat{y}_i)}{\partial \hat{y}_i} \dfrac{\partial \hat{y}_i}{\partial h_i} 
 \left( \prod_{i \geq j \geq k} \dfrac{\partial h_j}{\partial h'_{j}} \dfrac{\partial h'_j}{\partial h_{j-1}} \right) \leq \dfrac{\partial l(\hat{y}_i)}{\partial \hat{y}_i} \dfrac{\partial \hat{y}_i}{\partial h_i}\eta^{i-k}.
\end{equation}

As $\eta < 1$ it follows from \ref{eq:eta3} that in the presence of long-term sequences of data (for which $i-k$ is large) the product decreases exponentially fast to zero leading to the vanishing of the gradients.

\subsubsection{Exploding gradient}

Inverting the vanishing gradient proof, it is necessary that $\lambda \lambda' > \dfrac{1}{\gamma}$ for 

\begin{equation*}
\forall j, \left \lVert  \dfrac{\partial h_j}{\partial h'_j} diag\left(\dfrac{\partial h'_j}{\partial h_{j-1}}\right)  \right \rVert > \dfrac{1}{\gamma} \gamma > 1.
\end{equation*}

For all $j$, taking $\eta \in \mathbb{R}$ so that $\eta>1$ comes $\left \lVert  \dfrac{\partial h_j}{\partial h'_j}  diag\left(\dfrac{\partial h'_j}{\partial h_{j-1}}\right)  \right \rVert \geq \eta > 1$. By induction over $j$ we obtain

\begin{equation}
\label{eq:eta4}
    \dfrac{\partial l(\hat{y}_i)}{\partial \hat{y}_i} \dfrac{\partial \hat{y}_i}{\partial h_i} 
 \left( \prod_{i \geq j \geq k} \dfrac{\partial h_j}{\partial h'_{j}} \dfrac{\partial h'_j}{\partial h_{j-1}} \right) \geq \dfrac{\partial l(\hat{y}_i)}{\partial \hat{y}_i} \dfrac{\partial \hat{y}_i}{\partial h_i}\eta^{i-k}.
\end{equation}

As $\eta > 1$ it follows from \ref{eq:eta4} that in the presence of long-term sequences of data (for which $i-k$ is large) the product increases exponentially fast leading to the explosion of the gradients.

\subsection{Latent ODEs and the vanishing/exploding gradient problem} \label{proof:Latent ODE}

Let $l(\hat{y}_i)$ to be the contribution of the loss function at time step $i$ to the total loss $\mathcal{L}_{\Theta}$, where $\Theta=(\phi, \alpha)$ are the encoder and decoder parameters, respectively.
To find the parameters $\Theta$ that minimise the total loss, backpropagation is done:

\begin{equation}
\label{eq:latentODE1}
\dfrac{\partial l(\hat{y}_i)}{\partial \Theta} = \sum_{k=1}^i  \dfrac{\partial l(\hat{y}_i)}{\partial \hat{y}_i} \dfrac{\partial \hat{y}_i}{\partial \text{O}_{\text{NN}}} \dfrac{\partial \text{O}_{\text{NN}}}{\partial z_{t_i}} \dfrac{\partial z_{t_i}}{\partial z_0} \left(\dfrac{\partial z_0}{\partial \mu} \dfrac{\partial \mu}{\partial g} \dfrac{\partial g}{\partial h_i} + \dfrac{\partial z_0}{\partial \sigma} \dfrac{\partial \sigma}{\partial g} \dfrac{\partial g}{\partial h_i} \right) \dfrac{\partial h_i}{\partial h_k} \dfrac{\partial^+ h_k }{\partial \Theta}.
\end{equation}

\subsubsection{Vanishing gradient}

Due to the RNN encoder, the term $\dfrac{\partial h_i}{\partial h_k}$ has the same form as \eqref{eq:rnn3}, so, following the proof \ref{proof:RNN}, for all $j$, taking $\eta \in \mathbb{R}$ so that $\eta<1$ comes
$\forall j, \left \lVert  \dfrac{\partial h_j}{\partial h_{j-1}}  \right \rVert \leq \eta < 1$. By induction over $j$ we obtain,

\begin{multline}
\label{eq:eta4}
    \dfrac{\partial l(\hat{y}_i)}{\partial \hat{y}_i} \dfrac{\partial \hat{y}_i}{\partial \text{O}_{\text{NN}}} \dfrac{\partial \text{O}_{\text{NN}}}{\partial z_{t_i}} \dfrac{\partial z_{t_i}}{\partial z_0} \left(\dfrac{\partial z_0}{\partial \mu} \dfrac{\partial \mu}{\partial g} \dfrac{\partial g}{\partial h_i} + \dfrac{\partial z_0}{\partial \sigma} \dfrac{\partial \sigma}{\partial g} \dfrac{\partial g}{\partial h_i} \right) \left( \prod_{i \geq j \geq k} \dfrac{\partial h_j}{\partial h_{j-1}} \right) \leq \\
    \leq  \dfrac{\partial l(\hat{y}_i)}{\partial \hat{y}_i} \dfrac{\partial \hat{y}_i}{\partial \text{O}_{\text{NN}}} \dfrac{\partial \text{O}_{\text{NN}}}{\partial z_{t_i}} \dfrac{\partial z_{t_i}}{\partial z_0} \left(\dfrac{\partial z_0}{\partial \mu} \dfrac{\partial \mu}{\partial g} \dfrac{\partial g}{\partial h_i} + \dfrac{\partial z_0}{\partial \sigma} \dfrac{\partial \sigma}{\partial g} \dfrac{\partial g}{\partial h_i} \right)\eta^{i-k}.
\end{multline}

As $\eta < 1$ it follows from \ref{eq:eta4} that in the presence of long-term sequences of data (for which $i-k$ is large) the term $\eta^{i-k}$ decreases exponentially fast to zero leading to the vanishing of the gradients.

\subsubsection{Exploding gradient}

Similarly, for all $j$, taking $\eta \in \mathbb{R}$ so that $\eta>1$ comes
$\forall j, \left \lVert  \dfrac{\partial h_j}{\partial h_{j-1}}  \right \rVert \geq \eta > 1$. By induction over $j$ we obtain,

\begin{multline}
\label{eq:eta5}
    \dfrac{\partial l(\hat{y}_i)}{\partial \hat{y}_i} \dfrac{\partial \hat{y}_i}{\partial \text{O}_{\text{NN}}} \dfrac{\partial \text{O}_{\text{NN}}}{\partial z_{t_i}} \dfrac{\partial z_{t_i}}{\partial z_0} \left(\dfrac{\partial z_0}{\partial \mu} \dfrac{\partial \mu}{\partial g} \dfrac{\partial g}{\partial h_i} + \dfrac{\partial z_0}{\partial \sigma} \dfrac{\partial \sigma}{\partial g} \dfrac{\partial g}{\partial h_i} \right) \left( \prod_{i \geq j \geq k} \dfrac{\partial h_j}{\partial h_{j-1}} \right) \geq \\
    \geq  \dfrac{\partial l(\hat{y}_i)}{\partial \hat{y}_i} \dfrac{\partial \hat{y}_i}{\partial \text{O}_{\text{NN}}} \dfrac{\partial \text{O}_{\text{NN}}}{\partial z_{t_i}} \dfrac{\partial z_{t_i}}{\partial z_0} \left(\dfrac{\partial z_0}{\partial \mu} \dfrac{\partial \mu}{\partial g} \dfrac{\partial g}{\partial h_i} + \dfrac{\partial z_0}{\partial \sigma} \dfrac{\partial \sigma}{\partial g} \dfrac{\partial g}{\partial h_i} \right)\eta^{i-k}.
\end{multline}

As $\eta > 1$ it follows from \ref{eq:eta5} that in the presence of long-term sequences of data (for which $i-k$ is large) the term $\eta^{i-k}$ increases exponentially fast leading to the explosion of the gradients.

Thus, Latent ODEs suffer from the vanishing and exploding gradient problem due to the RNN used in the encoder.

\subsection{Latent ODE-RNNs and the vanishing/exploding gradient problem} \label{proof:Latent ODE-RNN}

Latent ODE-RNNs are Latent ODEs that have an ODE-RNN encoder instead of a RNN. 

The computation of the gradient of the loss at time step $i$, $\dfrac{\partial l(\hat{y}_i)}{\partial \Theta}$, is given by

\begin{equation}
\label{eq:latentODERNN1}
\dfrac{\partial l(\hat{y}_i)}{\partial \Theta} = \sum_{k=1}^i  \dfrac{\partial l(\hat{y}_i)}{\partial \hat{y}_i} \dfrac{\partial \hat{y}_i}{\partial \text{O}_{\text{NN}}} \dfrac{\partial \text{O}_{\text{NN}}}{\partial z_{t_i}} \dfrac{\partial z_{t_i}}{\partial z_0} \left(\dfrac{\partial z_0}{\partial \mu} \dfrac{\partial \mu}{\partial g_\text{NN}} \dfrac{\partial g_\text{NN}}{\partial z'_0} \dfrac{\partial z'_0}{\partial h_i} + \dfrac{\partial z_0}{\partial \sigma} \dfrac{\partial \sigma}{\partial g_\text{NN}} \dfrac{\partial g_\text{NN}}{\partial z'_0} \dfrac{\partial z'_0}{\partial h_i}  \right) \dfrac{\partial h_i }{\partial h_k} \dfrac{\partial^+ h_k}{\partial \Theta}.
\end{equation}

\subsubsection{Vanishing gradient}

Due to the ODE-RNN encoder, the term $\dfrac{\partial h_i}{\partial h_k}$ has the same form of \eqref{eq:odernn5} thus, it is sufficient to have $\lambda \lambda' < \dfrac{1}{\gamma}$ for 

 $$\forall j, \left \lVert  \dfrac{\partial h_j}{\partial h'_{j}} diag\left(\dfrac{\partial h'_j}{\partial h_{j-1}}\right)  \right \rVert < \dfrac{1}{\gamma} \gamma < 1$$.

 For all $j$, taking $\eta \in \mathbb{R}$ so that $\eta<1$ comes $\left \lVert  \dfrac{\partial h_j}{\partial h'_{j}} diag\left(\dfrac{\partial h'_j}{\partial h_{j-1}}\right)  \right \rVert \leq \eta < 1$. By induction over $j$ we obtain

\begin{multline}
\label{eq:eta6}
    \dfrac{\partial l(\hat{y}_i)}{\partial \hat{y}_i} \dfrac{\partial \hat{y}_i}{\partial \text{O}_{\text{NN}}} \dfrac{\partial \text{O}_{\text{NN}}}{\partial z_{t_i}} \dfrac{\partial z_{t_i}}{\partial z_0} \left(\dfrac{\partial z_0}{\partial \mu} \dfrac{\partial \mu}{\partial g_\text{NN}} \dfrac{\partial g_\text{NN}}{\partial z'_0} \dfrac{\partial z'_0}{\partial h_i} + \dfrac{\partial z_0}{\partial \sigma} \dfrac{\partial \sigma}{\partial g_\text{NN}} \dfrac{\partial g_\text{NN}}{\partial z'_0} \dfrac{\partial z'_0}{\partial h_i}  \right) \left( \prod_{i \geq j \geq k} \dfrac{\partial h_j}{\partial h'_{j}} \dfrac{\partial h'_j}{\partial h_{j-1}} \right) \leq \\
    \leq  \dfrac{\partial l(\hat{y}_i)}{\partial \hat{y}_i} \dfrac{\partial \hat{y}_i}{\partial \text{O}_{\text{NN}}} \dfrac{\partial \text{O}_{\text{NN}}}{\partial z_{t_i}} \dfrac{\partial z_{t_i}}{\partial z_0} \left(\dfrac{\partial z_0}{\partial \mu} \dfrac{\partial \mu}{\partial g_\text{NN}} \dfrac{\partial g_\text{NN}}{\partial z'_0} \dfrac{\partial z'_0}{\partial h_i} + \dfrac{\partial z_0}{\partial \sigma} \dfrac{\partial \sigma}{\partial g_\text{NN}} \dfrac{\partial g_\text{NN}}{\partial z'_0} \dfrac{\partial z'_0}{\partial h_i}  \right)\eta^{i-k}.
\end{multline}

As $\eta < 1$ it follows from \ref{eq:eta6} that in the presence of long-term sequences of data (for which $i-k$ is large) the term $\eta^{i-k}$ decreases exponentially fast to zero leading to the vanishing of the gradients.

\subsubsection{Exploding gradient}

Inverting the vanishing gradients proof, it is necessary to have $\lambda \lambda' > \dfrac{1}{\gamma}$ for 

 $$\forall j, \left \lVert  \dfrac{\partial h_j}{\partial h'_{j}} diag\left(\dfrac{\partial h'_j}{\partial h_{j-1}}\right)  \right \rVert > \dfrac{1}{\gamma} \gamma > 1$$.

 For all $j$, taking $\eta \in \mathbb{R}$ such that $\eta>1$ comes $\left \lVert  \dfrac{\partial h_j}{\partial h'_{j}} diag\left(\dfrac{\partial h'_j}{\partial h_{j-1}}\right)  \right \rVert \geq \eta > 1$. By induction over $j$ we obtain

\begin{multline}
\label{eq:eta7}
    \dfrac{\partial l(\hat{y}_i)}{\partial \hat{y}_i} \dfrac{\partial \hat{y}_i}{\partial \text{O}_{\text{NN}}} \dfrac{\partial \text{O}_{\text{NN}}}{\partial z_{t_i}} \dfrac{\partial z_{t_i}}{\partial z_0} \left(\dfrac{\partial z_0}{\partial \mu} \dfrac{\partial \mu}{\partial g_\text{NN}} \dfrac{\partial g_\text{NN}}{\partial z'_0} \dfrac{\partial z'_0}{\partial h_i} + \dfrac{\partial z_0}{\partial \sigma} \dfrac{\partial \sigma}{\partial g_\text{NN}} \dfrac{\partial g_\text{NN}}{\partial z'_0} \dfrac{\partial z'_0}{\partial h_i}  \right) \left( \prod_{i \geq j \geq k} \dfrac{\partial h_j}{\partial h'_{j}} \dfrac{\partial h'_j}{\partial h_{j-1}} \right) \geq \\
    \geq  \dfrac{\partial l(\hat{y}_i)}{\partial \hat{y}_i} \dfrac{\partial \hat{y}_i}{\partial \text{O}_{\text{NN}}} \dfrac{\partial \text{O}_{\text{NN}}}{\partial z_{t_i}} \dfrac{\partial z_{t_i}}{\partial z_0} \left(\dfrac{\partial z_0}{\partial \mu} \dfrac{\partial \mu}{\partial g_\text{NN}} \dfrac{\partial g_\text{NN}}{\partial z'_0} \dfrac{\partial z'_0}{\partial h_i} + \dfrac{\partial z_0}{\partial \sigma} \dfrac{\partial \sigma}{\partial g_\text{NN}} \dfrac{\partial g_\text{NN}}{\partial z'_0} \dfrac{\partial z'_0}{\partial h_i}  \right)\eta^{i-k}.
\end{multline}

As $\eta > 1$ it follows from \ref{eq:eta7} that in the presence of long-term sequences of data (for which $i-k$ is large) the term $\eta^{i-k}$ increases exponentially fast leading to the explosion of the gradients.

\end{document}